\documentclass[twoside]{article}

\usepackage[accepted]{aistats2024}
\usepackage[round]{natbib}

\usepackage{verbatim}
\usepackage{hyperref}
\hypersetup{
    colorlinks=true,
    linkcolor=blue,
    filecolor=magenta,      
    urlcolor=blue,
    citecolor=blue,
    }
\usepackage{wrapfig}
\usepackage{soul}
\usepackage{graphicx}
\usepackage{amsmath}
\usepackage{amssymb}
\usepackage{amsthm}
\usepackage{mathtools}
\usepackage{array}
\usepackage{bbm}
\usepackage{algorithm}
\usepackage{enumitem}
\usepackage{todonotes}
\usepackage{selectp}
\usepackage[algo2e]{algorithm2e}
\newcolumntype{x}[1]{>{\centering\let\newline\\\arraybackslash\hspace{0pt}}p{#1}}

\newcommand\bcolor[1]{\textcolor{blue}{\textbf{#1}}}

\newcolumntype{L}[1]{>{\raggedright\let\newline\\\arraybackslash\hspace{0pt}}m{#1}}
\newcolumntype{C}[1]{>{\centering\let\newline\\\arraybackslash\hspace{0pt}}m{#1}}
\newcolumntype{R}[1]{>{\raggedleft\let\newline\\\arraybackslash\hspace{0pt}}m{#1}}

\newtheorem{theorem}{Theorem}

\theoremstyle{definition}

\newtheorem{definition}{Definition}

\newcommand\numberthis{\addtocounter{equation}{1}\tag{\theequation}}

\begin{document}


\runningauthor{Davin Hill, Aria Masoomi, Max Torop, Sandesh Ghimire, Jennifer Dy}

\twocolumn[

\aistatstitle{Boundary-Aware Uncertainty for Feature Attribution Explainers}

\aistatsauthor{ Davin Hill \And Aria Masoomi \And Max Torop}
\aistatsaddress{Northeastern University \\ dhill@ece.neu.edu \And Northeastern University \\ masoomi.a@northeastern.edu  \And Northeastern University \\ torop.m@northeastern.edu}

\aistatsauthor{ Sandesh Ghimire \And Jennifer Dy}
\aistatsaddress{Northeastern University \\ drsandeshghimire@gmail.com \And Northeastern University \\ jdy@ece.neu.edu} ]

\begin{abstract}
%%%%%%%%%%%% Motivation
%As black-box classification models become increasingly relied upon in high-stakes applications, it becomes equally important to have reliable explanation methods.

%As many high-stakes applications have become increasingly dependant on black-box classification models, the development of reliable explanation methods has become crucial.

%precipitating a need for reliable explanations.
%Nevertheless, recent works have shown that many existing methods can be inconsistent or lack robustness.
Post-hoc explanation methods have become a critical tool for understanding black-box classifiers in high-stakes applications.
However, high-performing classifiers are often highly nonlinear and can exhibit complex behavior around the decision boundary, leading to brittle or misleading local explanations.
Therefore there is an impending need to quantify the uncertainty of such explanation methods in order to understand when explanations are trustworthy.
In this work we propose the \textbf{G}aussian \textbf{P}rocess \textbf{E}xplanation Un\textbf{C}ertainty (GPEC) framework, which generates a unified uncertainty estimate combining decision boundary-aware uncertainty with explanation function approximation uncertainty. We introduce a novel geodesic-based kernel, which captures the complexity of the target black-box decision boundary. We show theoretically that the proposed kernel similarity increases with decision boundary complexity.
The proposed framework is highly flexible; it can be used with any black-box classifier and feature attribution method.
Empirical results on multiple tabular and image datasets show that the GPEC uncertainty estimate improves understanding of explanations as compared to existing methods.
% In this work, we propose a novel geodesic-based kernel which captures the complexity of the target black-box decision boundary. We show theoretically that the proposed kernel similarity increases with decision boundary complexity.
% In addition, we introduce the \textbf{G}aussian \textbf{P}rocess \textbf{E}xplanation Un\textbf{C}ertainty (GPEC) framework, which generates a unified uncertainty estimate combining decision boundary-aware uncertainty with existing explanation uncertainty methods.
%We propose a novel method that quantifies uncertainty of explanation algorithms while being capturing information on the decision boundary of black box deep learning models. We introduce a geodesic-based similarity metric
%We combine this complexity-aware uncertainty with approximation uncertainty from \hl{the} explainer using a Gaussian Process model, which we call the Gaussian Process Explainer Uncertainty framework (GPEC).

\end{abstract}

% \vspace{-2mm}
\section{INTRODUCTION}
\vspace{-2mm}
%==========================================
% Intro
%==========================================
Post-hoc explainability methods have become a crucial tool for understanding and diagnosing their black-box model predictions. 
Recently, many such \emph{explainers} have been introduced in the category of local feature attribution methods; that is, methods that return a real-valued score representing each feature's relative importance for the model prediction.
These explainers are \emph{local} in that they are not limited to using the same decision rules throughout the data distribution, therefore they are better able to represent nonlinear and complex black-box models.

However, recent works have shown that local explainers can be inconsistent or unstable.
For example, explainers may yield highly dissimilar explanations for similar samples \citep{alvarez-melis2018,zulqarnain_lipschitzness}, exhibit sensitivity to imperceptible perturbations \citep{dombrowski2019explanations,ghorbaniInterpretationNeuralNetworks2019, slack_foolingLIMESHAP}, or lack stability under repeated application \citep{VSI_visani2022}.
When working in high-stakes applications, it is imperative to provide the user with an understanding of whether an explanation is reliable, potentially problematic, or even misleading.
A way to guide users regarding an explainer's reliability is to provide corresponding \emph{uncertainty quantification} estimates.

One can consider explainers as function approximators; as such, standard techniques for quantifying the uncertainty of estimators can be utilized to quantify the uncertainty of explainers. This is the strategy utilized by existing methods that estimate explainer uncertainty (e.g. \citep{reliable:neurips21,cxplain}). 
However, we observe that for explainers, this is not sufficient; in addition to uncertainty due to the {\em function approximation} of explainers, explainers also have to deal with the uncertainty due to the complexity of the {\em decision boundary (DB)} of the black-box model in the local region being explained.

\begin{figure}[t]
    \begin{center}
    \includegraphics[width=1\linewidth]{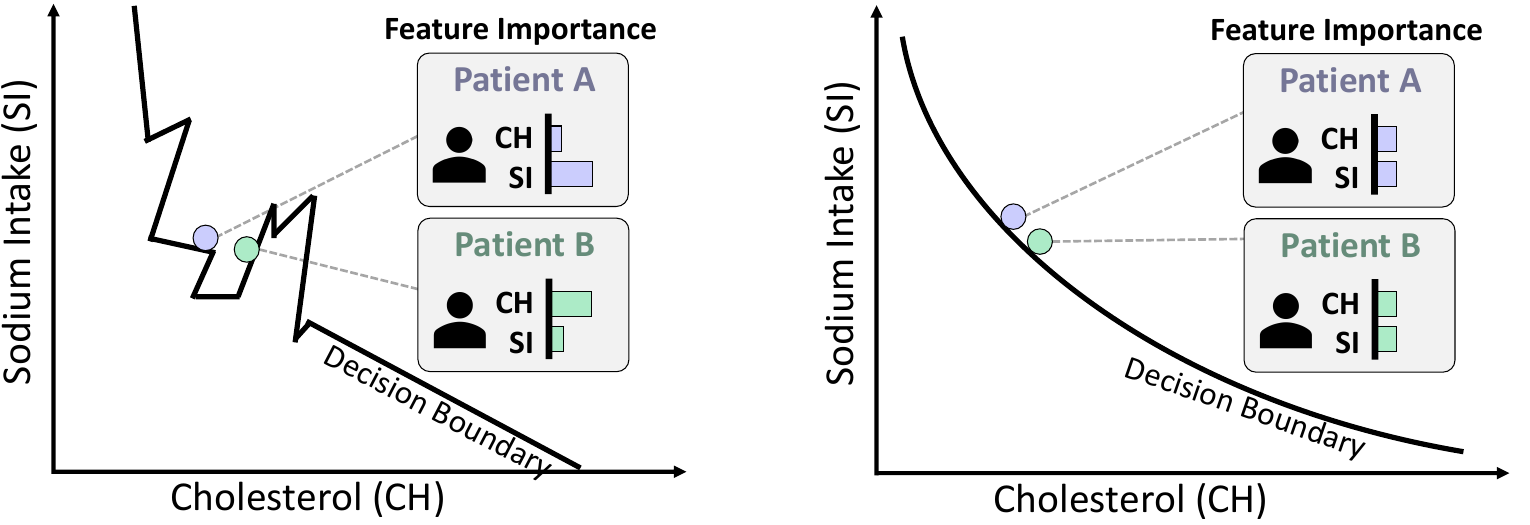}
    \end{center}
    \vspace{-2mm}
    \caption{Illustrative example of potential pitfalls when relying on local explainers for samples near complex regions of the decision boundary (left) as compared with a smoothed decision boundary (right).}
    \label{fig:example1}
    \vspace{-2mm}
\end{figure}

Previous works investigating DB geometry have related higher DB complexity to increased model generalization error \citep{valle-perez2018deep} and increased adversarial vulnerability \citep{moosavi2019robustness, Fawzi_topology_geometry_deep_networks}. Smoother DBs have been shown to improve feature attributions \citep{wang2020smoothed} and produce more consistent counterfactual explanations \citep{black2022consistent}.
\cite{dombrowski2019explanations} show that, in ReLU networks, samples with similar predictions can yield widely disparate explanations, which can be regulated through model smoothing.
Consider the following example (Fig. \ref{fig:example1}): a prediction model is used for a medical diagnosis using two features: cholesterol level and sodium intake.
We use the gradient with respect to each feature as an estimate of feature importance. Patients A and B have similar cholesterol and sodium levels and receive the same prediction, however, the complex decision boundary (left) results in a different top feature for each patient. In contrast, the smoothed decision boundary (right) yields more consistent explanations.

\begin{figure*}[t]
    \begin{center}
    \includegraphics[width=0.95\linewidth]{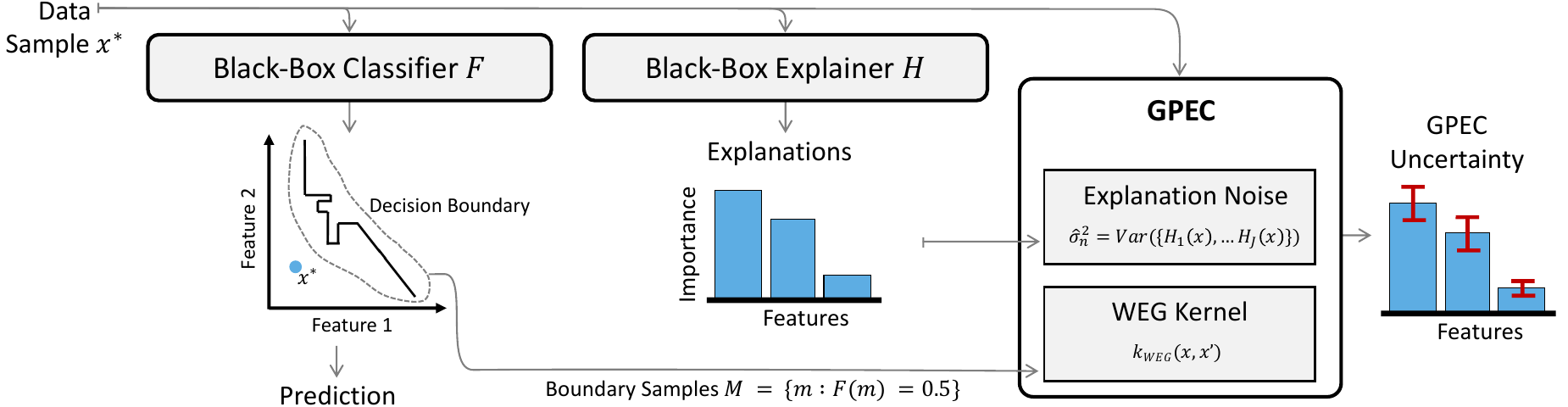}
    \end{center}
    \vspace{-2mm}
    \caption{Overview of the GPEC framework. GPEC takes samples from the classifier's decision boundary plus (possibly noisy) explanations and fits a GP model with the novel WEG Kernel. The GPEC estimate incorporates both the uncertainty derived from the decision boundary complexity and also the explanation approximation uncertainty from the explainer.}
    \label{fig:fig1}
    \vspace{-2mm}
\end{figure*}

%==========================================
% Introduce our method.
%==========================================
We approach this problem from the perspective of similarity: given two samples and their respective explanations, how closely related should the explanations be? From the previous intuition, we define this similarity based on a geometric perspective of the DB complexity between these two points.
Specifically, we propose the novel Weighted Exponential Geodesic (WEG) kernel, which encodes our expectation that two samples close in Euclidean space may not actually be similar if the DB within a local neighborhood of the samples is highly complex.

Using this similarity formulation, we propose the \textbf{G}aussian \textbf{P}rocess \textbf{E}xplanation Un\textbf{C}ertainty (GPEC) framework (Fig. \ref{fig:fig1}), which is an instance-wise, model-agnostic, and explainer-agnostic method to quantify the explanation uncertainty.
The proposed notion of uncertainty is complementary to existing quantification methods. Existing methods primarily estimate the uncertainty related to the choice in model parameters and fitting the explainer, which we call \emph{function approximation uncertainty}, and does not capture uncertainty related to DB complexity.
GPEC can combine the DB-based uncertainty with function approximation uncertainty derived from any local feature attribution method.

%==========================================
% Summary
%==========================================
In summary, we make the following contributions:
\setlist{nolistsep}
\begin{itemize} [leftmargin=0.3cm]
    \itemsep0.4em 
    \item We introduce a novel geometric perspective on capturing explanation uncertainty and define a geodesic-based similarity between explanations. We prove theoretically that the proposed similarity captures the complexity of the decision boundary from a given black-box classifier.
    \item We propose a novel Gaussian Process-based framework that combines 1) uncertainty from decision boundary complexity and 2) explainer-specific function approximation uncertainty to generate uncertainty estimates for any given feature attribution method and black box model.
    \item Empirical results show GPEC uncertainty improves understanding of feature attribution methods.
\end{itemize}

\vspace{-1mm}
\section{RELATED WORKS} \label{sec:related_works}
\vspace{-2mm}
\textbf{Explanation Methods.}
A variety of methods have been proposed for improving the transparency of pre-trained black-box prediction models \citep{guidotti_survey, arrietta_survey}.
Within this category of \emph{post-hoc} methods, many methods focus on \emph{local} explanations, that is, explaining individual predictions rather than the entire model.
Some of these methods implement local \emph{feature selection} \citep{l2x_chen, IFG}; others return a real-valued score for each feature, termed \emph{feature attribution} methods, which are the primary focus of this work. For example, LIME \citep{ribeiroLIME} trains a local linear regression model to approximate the black-box model.
\citet{shap_lundberg} generalizes LIME and five other feature attribution methods using the SHAP framework, which fulfill a number of desirable axioms.
While LIME and SHAP are model-agnostic, others are model-specific, such as neural networks \citep{LRP_saliency, pmlr-v70-shrikumar17a, sundararajanAxiomaticAttributionDeep2017, erionImprovingPerformanceDeep2021}, tree ensembles \citep{lundbergLocalExplanationsGlobal2020}, or Bayesian neural networks \citep{bykov2020much}. Another class of methods involves training surrogate models to explain the black-box model \citep{dabkowski_gal, l2x_chen, cxplain, guo_explaining_nonparametric_approach, jethani2022fastshap}.

\textbf{Explanation Uncertainty.}
One option for improving explainer trustworthiness is to quantify their associated uncertainty. Bootstrap resampling techniques have been proposed to estimate uncertainty from surrogate-based explainers \citep{cxplain, schulz2022uncertainty}. \citet{guo_explaining_nonparametric_approach} also proposes a surrogate explainer parameterized with a Bayesian mixture model.
Alternatively, \citet{bykov2020much} and \citet{ucam_uncertainty} introduce methods for explaining Bayesian neural networks, which can be transferred to their non-Bayesian counterparts.
\citet{improving_kernelshap} derive an unbiased version of KernelSHAP and investigates an efficient way of estimating its uncertainty.
\citet{uncertainty_in_lime} categorizes different sources of variance in LIME estimates.
Several methods also investigate LIME and KernelSHAP in a Bayesian context; for example, calculating a posterior over attributions \citep{reliable:neurips21}, investigating priors for explanations \citep{bayelime_zhao21}, or using active learning during sampling \citep{select_wisely_and_explain_Saina2022}.

However, existing methods for quantifying explanation uncertainty only consider the uncertainty of the explainer as a function approximator.
This work introduces an additional notion of uncertainty for explainers that considers the complexity of the classifier DB.
\vspace{-2mm}
\section{UNCERTAINTY FRAMEWORK FOR EXPLAINERS} \label{sec:GP_basic}
\vspace{-2mm}
\begin{comment}
\end{comment}
%==========================================
% Why GP?
%==========================================
We now outline the GPEC framework (Fig. \ref{fig:fig1}), which is parametrized with a Gaussian Process (GP) regression model\footnote{A brief review of GP regression is provided in App. \ref{app:background}.}.
%==========================================
% Concrete formulation of GP
%==========================================
Consider a sample $x^* \in \mathcal{X}$ that we want to explain in the context of a black-box classifier $F:\mathcal{X} \to [0,1]$, where $\mathcal{X} \subseteq \mathbb{R}^D$ is the data space and $D$ is the number of features.
For convenience we consider the binary classification case; this is extended to multiclass in App. \ref{app:proof}. We apply a local feature attribution explainer $H:\mathcal{X} \to \mathbb{R}^D$.

Recent works (e.g. \citet{alvarez-melis2018, dombrowski2019explanations}) have shown that local explanations can lack robustness and stability related to model complexity.
Therefore, when explaining samples in high-stakes applications, it is critical to understand the behavior of the explainer, especially in relation to other samples near $x^*$.
More concretely, let $X\in \mathbb{R}^{N \times D}$ represent a dataset of $N$ samples. Here, each row vector $X_n \in \mathbb{R}^D$ , $n \in N$ represents a data point. We apply $H$ to the rows of $X$ generating $N$ observed explanations, $E_n \in \mathbb{R}^D$ , $n \in N$, which are grouped into $E \in \mathbb{R}^{N \times D}$.
We can use these observed sample-explanation pairs to infer the behavior of $H$ around $x^*$, however there are two main challenges.
First, we expect the similarity between the explanations of $X$ and $x^*$ to be dependent on $F$.
In particular, we expect that as the DB in a neighborhood around $x^*$ and a given sample $X_n$ becomes increasingly complex, $H(x^*)$ and $H(X_n)$ may become more dissimilar; i.e. $H(X_n)$ may not contain useful information towards inferring $H(x^*)$. In this situation, the user should be prompted to either draw additional samples near $x^*$, or otherwise be warned of higher uncertainty.
Second, the observed explanations $E$ can be noisy; many explainers are stochastic and approximated with sampling methods or a learned function.
%============

To solve these challenges, we can model the explainer with a vector-valued GP regression by treating the explainer as a latent function inferred using samples $X$ and explanations $E$.
We model each explanation $E_n$ as being generated from a latent function $\mathcal{H}$ plus independent Gaussian noise $\eta_n$. For convenience, we consider each feature $d$ independently; see App. \ref{app:proof} for extensions.

\vspace{-15pt}
% Let $\{x_1, \dots, x_N\}$ represent a set of observed samples with respective explanations $\{e_1, \dots, e_N\}$.
% We assume that each explanation $e_n$ is generated from a latent function $\mathcal{H}$ plus independent Gaussian noise $\eta_n$.
% \begin{equation} \label{eqn:GP}
% % \setlength{\abovedisplayskip}{10pt}
% % \setlength{\belowdisplayskip}{3pt}
%    e_n = \mathcal{H}(x_n) + \eta_n  \quad s.t. \quad  \underbrace{\mathcal{H}(x_n) \sim \mathcal{GP}(0,k(x, x'))}_{\textrm{Decision Boundary-Aware Uncertainty}} \quad \underbrace{\eta_n \sim \mathcal{N}(0,\sigma_n^{2}  )}_{\textrm{Function Approximation Uncertainty}} 
% \end{equation}
\small \begin{align*} \numberthis \label{eqn:GP}
   E_{n,d} = \mathcal{H}_d(X_n) + \eta_{n,d}  \; \; s.t.  \underbrace{\mathcal{H}_d(X_n) \sim \mathcal{GP}(0,k(\cdot, \cdot))}_{\textrm{Decision Boundary-Aware Uncertainty}} \\
   s.t. \underbrace{\eta_{n,d} \sim \mathcal{N}(0,\sigma_{n,d}^{2}  )}_{\textrm{Function Approximation Uncertainty}} 
\end{align*}
\normalsize
where $k(\cdot, \cdot)$ is the specified kernel function for the GP prior.
% How does GP capture two sources of uncertainty?
We disentangle each explanation into two components, $\mathcal{H}(X_n)$ and $\eta_n$, which represent two separate sources of uncertainty: 1) a \emph{decision boundary-aware }uncertainty which we capture using the kernel similarity, and 2) a \emph{function approximation} uncertainty from the explainer.
After specifying $\mathcal{H}(X_n)$ and $\eta_n$, we can combine the two sources by calculating the predictive distribution for $x^*$. We take the variance of this distribution as the GPEC uncertainty estimate:
% \begin{align*} \numberthis \label{eqn:GP_variance}
% \mathbb{V}[x^*] = k(x^*, x^*) - k(x^*, x)[K + \sigma^2 I_N]^{-1} k(x, x^*)
% \end{align*}
\begin{equation}  \label{eqn:GP_variance}
\mathbb{V}_d[x^*] = k(x^*, x^*) -  k(X, x^*)^\intercal[K + \sigma_{d}^2 I_N]^{-1} k(X, x^*)
\end{equation}
where $K \in \mathbb{R}^{N \times N}$ is the kernel matrix s.t. $K_{ij} = k(X_i, X_j)$ $\forall i,j \in \{1...N\}$, $k(X, x^*) \in \mathbb{R}^{N \times 1}$ has elements $k(X,x^*)_i = k(X_i, x^*)$ $i \in \{1...N\}$ , $\sigma_d^2 \in \mathbb{R}^{N}_{+}$ is the variance parameter for explanation noise, and $I_N$ is the identity matrix. From Eq. \eqref{eqn:GP_variance} we see that predictive variance captures DB-aware uncertainty through the kernel function $k(\cdot, \cdot)$, and also the function approximation uncertainty through the $\sigma^2_{d} I_N$ term.

%==========================================
% Functional Approximation Uncertainty
%==========================================
\textbf{Function Approximation Uncertainty.} The $\eta_n$ component of Eq. \eqref{eqn:GP} represents the uncertainty stemming from explainer estimation. 
For example, $\eta_n$ can represent the variance due to sampling (e.g. perturbation-based explainers) or explainer training (e.g. surrogate-based explainers).
Explainers that include some estimate of uncertainty (e.g. BayesLIME, BayesSHAP, CXPlain) can be directly used to estimate $\sigma^2_n$.
For other stochastic explanation methods, we can estimate $\sigma^2_n$ empirically by resampling $J$ explanations for the same sample $X_n$:
% \begin{equation} \label{eqn:empirical_uncertainty}
%    \hat \sigma_n^{2} = [\frac{1}{|J|}\sum_{i=1}^{J} (E_i(x_{n}) - \bar E(x_n))^2]^{-1} \quad s.t.  \quad \bar E(x_n) = \frac{1}{|J|} \sum_{i=1}^J E_i(x_{n})
% \end{equation}
\begin{align*} \numberthis \label{eqn:empirical_uncertainty}
   \hat \sigma_n^{2} = \frac{1}{|J|}\sum_{i=1}^{J} \big(H_i(X_{n}) - \frac{1}{|J|} \sum_{j=1}^J H_j(X_{n}) \big)^2
\end{align*}
% Two line Equation
% \begin{align*} \numberthis \label{eqn:empirical_uncertainty}
%    \hat \sigma_n^{2} = [\frac{1}{|J|}\sum_{i=1}^{J} (E_i(x_{n}) - \bar E(x_n))^2]^{-1} \\
%    s.t. \quad \bar E(x_n) = \frac{1}{|J|} \sum_{i=1}^J E_i(x_{n})
% \end{align*}
where each $H_i(X_n)$ is a sampled explanation. Alternatively, for deterministic explanation methods we can omit the $\eta_n$ term and assume noiseless explanations.

%==========================================
% Decision Boundary Uncertainty
%==========================================
\textbf{Decision Boundary-Aware Uncertainty.}
% In contrast, the $\mathcal{H}(X_n)$ component of Eq. \eqref{eqn:GP} draws possible functions from the GP prior that could have generated the observed explanations; the distribution of these functions serves as an estimate of uncertainty.
In contrast, the $\mathcal{H}(X_n)$ component of Eq. \eqref{eqn:GP} represents the distribution of functions that could have generated the observed explanations.
The choice of kernel $k(\cdot, \cdot)$ encodes our \emph{a priori} assumption regarding the similarity between explanations based on the similarity of their corresponding inputs.
In other words, given two samples $x, x' \in \mathcal{X}$, how much information do we expect a given explanation $H(x)$ to provide for a nearby explanation $H(x')$?
As the DB between $H(x)$ and $H(x')$ becomes more complex, we would expect for this information to decrease.
In Section \ref{sec:weighted_geodesic_kernel}, we consider a novel kernel formulation that reflects the complexity of the DB in a local neighborhood of the samples.

%%%%%%%%%%%%%%%%%%%%%%%%%%%%%%%%%%%%%%%%%%%%%%%%%%%%%%%%%%%%%%%%%%%%%%
%%%%%%%%%%%%%%%%%%%%%%%%%%%%%%%%%%%%%%%%%%%%%%%%%%%%%%%%%%%%%%%%%%%%%%
%%%%%%%%%%%%%%%%%%%%%%%%%%%%%%%%%%%%%%%%%%%%%%%%%%%%%%%%%%%%%%%%%%%%%%
\vspace{-1mm}
\section{WEG KERNEL} \label{sec:weighted_geodesic_kernel}
\vspace{-2mm}
\begin{figure*}[t]
   \setlength{\abovecaptionskip}{-5pt}
   \begin{center}
   \includegraphics[width= 0.95\linewidth]{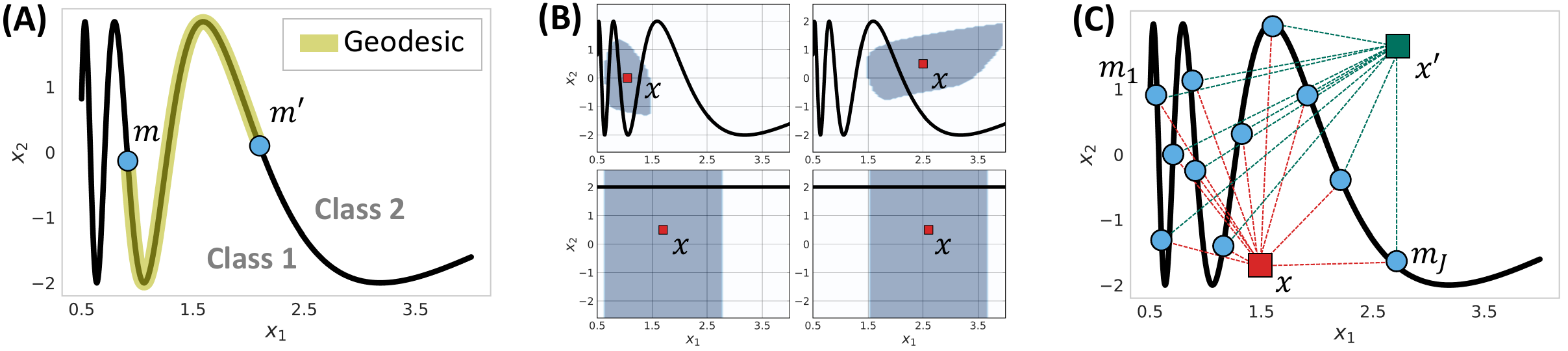}
   \end{center}
   \caption{Consider a classifier with DB defined as $\mathcal{M}_{0} = \{\left(x_1, f(x_1)\right): x_1 \in \mathbb{R}_{>0}\}$ where $f(x_1) = 2\cos(\frac{10}{x_1})$. \textbf{(A)} Illustration of geodesic distance $d_{geo}(m, m')$ between two points $m', m \in \mathcal{M}_{0}$.
   \textbf{(B)} Evaluation of the WEG kernel for $\mathcal{M}_{0}$ (top) and a linear DB (below). The gray region highlights the set $\{x' : k(x,x') \geq 0.9\}$ for a given $x$ (red). This region increases as the local DB become more linear.
   \textbf{(C)} During WEG approximation, we calculate Euclidean distances between $x,x'$ (red, green) and DB samples $m_1,...,m_J \in \mathcal{M}_{0}$ (blue). When appropriately normalized (Eq. \eqref{eq:q_weighting}), this acts as a weighting for each element of the EG kernel.}
   \label{fig:kernel_comparison}
\end{figure*}
%==========================================
% Motivate Decision Boundary Awareness
%==========================================

%==========================================
% How is similarity captured in the GP?
%==========================================
\begin{comment}
Given a sample $x$, kernel $k(x,x')$ and $\alpha\in (0,1)$, we can define this \emph{explanation neighborhood} as:
\begin{equation}
\mathcal{N}_{\alpha}(x) =\{x' : k(x,x') \geq \alpha \}
\end{equation}
\end{comment}
Intuitively, the GP kernel encodes the assumption that each explanation provides some information about other nearby explanations, which is defined through kernel similarity.
% In GPEC, the kernel defines the similarity between two explanations $\mathcal{H}(x)$ and $\mathcal{H}(x')$ based on some function of their inputs $x$ and $x'$.
To capture boundary-aware uncertainty, we want to define a similarity $k(x,x')$ that is inversely related to the complexity or smoothness of the DB between $x,x'\in \mathcal{X}$.

\vspace{-4pt}
\subsection{Geometry of the Decision Boundary}
\vspace{-3pt}
%==========================================
% Why Geodesic?
%==========================================
% \begin{wrapfigure}{r}{0.66\textwidth}
%    \begin{center}
%       \includegraphics[width= 1\linewidth]{figures/figure_geodesic.pdf}
%       \caption{Consider a classifier with DB defined as $f_{\alpha}(x) = 2\sin(\frac{\alpha}{x})$. LEFT: Comparison of geodesic and Euclidean distance between $f_{\alpha}(0.5)$ and $f_{\alpha}(4)$ for varying $\alpha$. RIGHT: Evaluation of the WEG kernel for different points (red) on $f_{\alpha}$. The gray region highlights the set $\{x' : k(x,x') \geq 0.9\}$ for the given red point $x$. This region increases as nearby sections of $f_\alpha$ become more linear.}
% %    \setlength{\belowcaptionskip}{-5pt}
%    \label{fig:kernel_comparison}
%    \end{center}
% \end{wrapfigure}
We represent the DB as a hypersurface embedded in $\mathbb{R}^D$ with co-dimension one. Given the classifier $F$, we define the DB\footnote{Without loss of generality, we assume that the classifier decision rule is $\frac{1}{2}$} as $\mathcal{M}_F = \{m \in \mathbb{R}^{D} : F(m) = \frac{1}{2} \}$.
For any two points $m, m' \in \mathcal{M}_F$, let $\gamma:[0,1] \rightarrow \mathcal{M}_F$ be a differentiable map such that $\gamma(0) = m$ and $\gamma(1) = m'$, representing a 1-dimensional curve on $\mathcal{M}_F$.
We can then define distances along the DB as geodesic distances in $\mathcal{M}_F$ (Fig \ref{fig:kernel_comparison}A):
% We can then define geodesic distances along the DB as the minimal length curve connecting $m,m'$ (Fig \ref{fig:kernel_comparison}A):
\begin{equation}
\setlength{\abovedisplayskip}{3pt}
\setlength{\belowdisplayskip}{3pt}
   d_{geo}(m,m') = \min_{\gamma} \int_{0}^{1} || \dot \gamma(t)|| \textrm{d}t \quad \;  \forall m,m' \in \mathcal{M}_F
\end{equation}
The relative complexity of the DB can be characterized by the geodesic distances between points on the DB.
% We want to relate kernel similarity to the behavior of the black-box model; specifically, the complexity or smoothness of the DB. Given any two points on the DB, the relative complexity of the boundary segment between them can be approximated by the segment length.
For example, the simplest form that the DB can take is a linear boundary. Consider a black-box model with linear DB $\mathcal{M}_1$. For two points $z,z' \in \mathcal{M}_1$, $d_{geo}(z,z') = ||z-z'||_2$ which corresponds with the minimum geodesic distance in the ambient space. For any nonlinear DB $\mathcal{M}_2$ that also contains $z,z'$, it follows that $d_{geo}(z,z') > ||z-z'||_2$. As the complexity of the DB increases, there is a general corresponding increase in geodesic distances between fixed points on the DB.
%==========================================
% Exponential Geodesic Kernel
%==========================================
We can adapt geodesic distance in our kernel selection through the exponential geodesic (EG) kernel \citep{feragen_geodesic_exponential_kernel}.
%, which is a generalization of the Radial Basis Function (RBF) kernel substituting $\ell_2$ distance with geodesic distance:
\begin{equation} \label{eqn:eg}
   k_{\textrm{EG}}(m, m') = \exp\left[-\lambda  d_{geo}(m,m')\right]
\end{equation}
%==========================================
% Problems of Exponential Geodesic Kernel (validity, points on manifold only)
%==========================================
The EG kernel has been previously investigated in the context of Riemannian manifolds \citep{feragen_geodesic_exponential_kernel, feragen_open_problem_geodesic_kernel}. In particular, while prior work shows that the EG kernel fails to be positive definite for all values of $\lambda$ in non-Euclidean space, there exists large intervals of $\lambda>0$ for which the EG kernel is positive definite. Appropriate values can be selected through grid search and cross validation; we assume that a valid value of $\lambda$ has been selected.

%==========================================
% EG to WEG
%==========================================
Therefore, by sampling $\mathcal{M}_F$, we can use the EG kernel matrix to capture DB complexity. However, a challenge remains in relating points $x, x' \in \mathcal{X} \setminus \mathcal{M}_F$ to the nearby DB. In Section \ref{sec:weighting} we consider a continuous weighting over $\mathcal{M}_F$ based on distance to $x,x'$.

% \vspace{-3pt}
\subsection{Weighting Decision Boundary Samples} \label{sec:weighting}
\vspace{-1mm}

% % \vspace{-2pt}
% \begin{figure}[t]
%    \begin{center}
%    \includegraphics[width= 0.7\linewidth]{figures/figure_geodesic.pdf}
%    \end{center}
% %    \setlength{\belowcaptionskip}{-5pt}
%    \caption{Consider a classifier with DB defined as $f_{\alpha}(x) = 2\sin(\frac{\alpha}{x})$. LEFT: Comparison of geodesic and Euclidean distance between $f_{\alpha}(0.5)$ and $f_{\alpha}(4)$ for varying $\alpha$. RIGHT: Evaluation of the WEG kernel for different points (red) on $f_{\alpha}$. The gray region highlights the set $\{x' : k(x,x') \geq 0.9\}$ for the given red point $x$. This region increases as nearby sections of $f_\alpha$ become more linear.}
% \end{figure}

%==========================================
% Weighting
%==========================================
Let $p(m)$ denote a distribution with support defined over $\mathcal{M}_F$ such that we can draw DB samples $m_1...m_J \sim p(m)$ using a DB sampling algorithm (see Sec. \ref{sec:alg}).
%For example, we can define $p(M)$ to be uniform over the decision boundary
% what does this distribution look like?
We weight $p(m)$ according to the $\ell_2$ norm between $m$ and a fixed data sample $x$ to create a weighted distribution $q( m | x, \rho)$:
\begin{equation} \label{eq:q_weighting}
    q(m | x, \rho) \propto  \exp\left[-\rho || x-m||_2^2\right] p(m) 
\end{equation}
where $\rho$ represents a hyperpameter that controls the sensitivity of the weighting. We can then define the kernel function $k_{\textrm{WEG}}(x,x')$ by taking the expected value over the weighted distributions.
% \begin{equation} \label{eq:kernel_weight} 
% k_{\textrm{WEG}}(x,x') = \int \int \exp[-\lambda d_{geo}(m, m')] \; q(m|x, \rho) \; q(m'|x', \rho) \; \textrm{d} m \textrm{d} m'
% \end{equation}
% \begin{align*} \label{eq:kernel_weight1} 
% k_{\textrm{WEG}}(x,x') = \int \int \exp\left[-\lambda d_{geo}(m, m')\right] \quad \quad  \\
% \times \; q(m|x, \rho) \; q(m'|x', \rho) \; \textrm{d} m \textrm{d} m'
% \numberthis 
% \end{align*}
\begin{align*} \label{eq:kernel_weight1} 
k_{\textrm{WEG}}(x,x') = \int \int k_{\textrm{EG}}(m, m') \quad \quad \quad \quad \quad \quad  \\
\times \; q(m|x, \rho) \; q(m'|x', \rho) \; \textrm{d} m \textrm{d} m'
\numberthis 
\end{align*}
% \begin{align*} \label{eq:kernel_weight} 
% k_{\textrm{WEG}}(x,x') = \mathbb{E}_{m \sim  q(m|x, \rho), m' \sim q(m'|x', \rho) }\left[k_{EG}(m,m')\right]
% \numberthis 
% \end{align*}
% \begin{equation}\label{eq:kernel_weight1} 
% k_{\textrm{WEG}}(x,x') = \int \int k_{\textrm{EG}}(m, m') \; q(m|x, \rho) \; q(m'|x', \rho) \; \textrm{d} m \textrm{d} m'
% \end{equation}
% % \begin{equation} \label{eq:kernel_weight} 
% % = \frac{\int \int \exp[-\lambda d_{geo}(m, m') -\rho (|| x-m||_2^2 + || x-m'||_2^2)] p(m) p(m')  \; \textrm{d} m \textrm{d} m'}{(\int \exp[-\rho || x-m||_2^2] p(m) \; \textrm{d}m)  (\int \exp[-\rho || x-m'||_2^2] p(m') \; \textrm{d}m') }
% % \end{equation}
\begin{align*} \label{eq:kernel_weight} 
   = \frac{1}{Z_m Z_{m'}}\int \int \exp\left[-\lambda d_{geo}(m, m') \right]  \quad \\
\times \exp\left[-\rho (|| x-m||_2^2 + || x'-m'||_2^2)\right] p(m) p(m')  \textrm{d} m \textrm{d} m'
\numberthis
\end{align*}
%==========================================
% Kernel Validity
%==========================================
where $Z_m, Z_{m'}$ are normalizing constants for $q(m | x, \rho)$ and $q(m' | x', \rho)$, respectively. Eq. \eqref{eq:kernel_weight} is an example of a marginalized kernel \citep{tsuda_marginalizedkernel}: a kernel defined by the expected value of observed samples $x,x'$ over latent variables $m,m'$. Given that the underlying EG kernel is positive definite, it follows that the WEG kernel forms a valid kernel.

With the WEG kernel, we can calculate a similarity between $x,x'\in \mathcal{X}$ that decreases as the complexity of the DB segments between the two points increases.
In Fig. \ref{fig:kernel_comparison}B we evaluate the WEG kernel similarity on nonlinear and a linear DB.  We observe that WEG similarity reflects the complexity of the DB; as the decision boundary becomes more linear in a local region, the similarity between neighboring points increases.
% introduce theorems
To evaluate the WEG kernel theoretically, we consider two properties. Theorem \ref{thm:kernel_expgeodesic} establishes that the EG kernel is a special case of the WEG kernel for when $x,x' \in \mathcal{X} \cap \mathcal{M}_F$.

%==========================================
% Theorem: Equivalence to Exponential Geodesic
%==========================================
\vspace{4pt}
\begin{theorem} \label{thm:kernel_expgeodesic}
   Given two points $x,x' \in \mathcal{X} \cap \mathcal{M}_F$, then $\lim_{\rho \rightarrow \infty} k_{\textrm{WEG}}(x,x') = k_{\textrm{EG}}(x, x')$
\end{theorem}
Proof details are shown in App. \ref{proof:points_on_manifold}. Intuitively, as $\rho$ increases the manifold distribution closest to the points $x,x'$ becomes weighted increasingly heavily. At the limit, the weighting concentrates entirely on $x,x'$ themselves, which recovers the EG kernel. Therefore we see that the WEG kernel is a generalization of the EG kernel with a weighting controlled by $\rho$.  

%==========================================
% Theorem: Kernel Complexity
%==========================================
Theorem \ref{thm:kernel_complexity} establishes the inverse relationship between DB complexity and WEG similarity. Given a classifier with a piecewise linear DB, we show that this DB represents a local maximum with respect to WEG kernel similarity; i.e. as we perturb the DB to be nonlinear, kernel similarity decreases.
We first define \emph{perturbations} on the DB. Note that $\textrm{int}(S)$ indicates the interior of a set $S$ and $\textrm{id}$ indicates the identity mapping.  

\vspace{4pt}

\begin{definition}[Manifold Perturbation] \label{def:1}
Let $\{U_\alpha\}_{\alpha\in I}$ be charts of an atlas for a manifold $\mathcal{P} \subset \mathbb{R}^D$, where $I$ is a set of indices. Let $\mathcal{P}$ and $\widetilde{\mathcal{P}}$ be differentiable manifolds embedded in $\mathbb{R}^D$, where $\mathcal{P}$ is a Piecewise Linear manifold. Let  $R: \mathcal{P} \rightarrow \widetilde{\mathcal{P}}$ be a diffeomorphism. We say $\widetilde{\mathcal{P}}$ is a \emph{perturbation} of $\mathcal{P}$ on the $i^{\textrm{th}}$ chart if $R$ satisfies the following two conditions:
\raisebox{.5pt}{\textcircled{\raisebox{-.9pt} {1}}} There exists a compact subset $K_i \subset U_i$ s.t.  $\left.R\right|_{\mathcal{P} \setminus \textrm{int}(K_i)} = \left.\textrm{id}\right|_{\mathcal{P} \setminus \textrm{int} (K_i)}$ and $\left.R\right|_{\textrm{int}(K_i)} \neq \left.\textrm{id}\right|_{\textrm{int}(K_i)}$.
\raisebox{.5pt}{\textcircled{\raisebox{-.9pt} {2}}} There exists a linear homeomorphism between an open subset $\widetilde{U_i} \subseteq U_i$ with $\mathbb{R}^{d-1}$ which contains $K_i$.
\end{definition}

\vspace{4pt}

\begin{theorem} \label{thm:kernel_complexity}
Let $\mathcal{P}$ be a $(d$--$1)$-dimension Piecewise Linear manifold embedded in $\mathbb{R}^D$. Let $\widetilde{\mathcal{P}}$ be a perturbation of $\mathcal{P}$ and define $\tilde k(x,x')$ and $k(x,x')$ as the WEG kernels defined on $\widetilde{\mathcal{P}}$ and $\mathcal{P}$ respectively. Then $\tilde k(x,x') < k(x,x') \; \forall x,x' \in \mathbb{R}^D$.
\end{theorem}

Proof details are shown in App. \ref{proof:kernel_complexity}.
Theorem \ref{thm:kernel_complexity} implies that, for any two fixed points $x,x'$, their kernel similarity $k_{\textrm{WEG}}(x,x')$ decreases as the black-box DB complexity increases. Within GPEC, the explanations for $x,x'$ become less informative for other nearby explanations and induce a higher explanation uncertainty estimate.

%==========================================
% Normalization
%==========================================
To improve the WEG kernel interpretation, we can normalize $k_{\textrm{WEG}}$ to scale similarity values to be between $[0,1]$. We define the normalized kernel $k_{\textrm{WEG}}^*$:
\begin{equation}
\setlength{\belowdisplayskip}{-2pt}
    k_{\textrm{WEG}}^*(x,x') = \frac{k_{\textrm{WEG}}(x,x')}{\sqrt{k_{\textrm{WEG}}(x,x) k_{\textrm{WEG}}(x',x')}}
\end{equation}
% \vspace{-7pt}
% \vspace{-3pt}
\subsection{WEG Kernel Approximation}
\vspace{-1mm}
In practice, the integrals in Eq. \eqref{eq:kernel_weight} are intractable; we can instead use Monte Carlo integration to approximate $k_{\textrm{WEG}}(x,x')$ with $J$ samples $m_1,..., m_J \sim p(m)$.
% \begin{equation} \label{eq:approximation}
%     k_{\textrm{WEG}}(x,x') \approx \frac{1}{Z_m Z_{m'} J^2} \sum_{i=1}^J \sum_{j=1}^J \underbrace{\exp[-\lambda d_{geo}(m_i, m_j)]}_{k_{\textrm{EG}}(m_i, m_j)} \exp[-\rho (|| x-m_i||_2^2 + || x'-m_j||_2^2)]
% \end{equation}
% \begin{align*} 
%     k_{\textrm{WEG}}(x,x') \approx \frac{1}{Z_m Z_{m'} J^2} \sum_{i=1}^J \sum_{j=1}^J \underbrace{\exp[-\lambda d_{geo}(m_i, m_j)]}_{k_{\textrm{EG}}(m_i, m_j)} \exp[-\rho (|| x-m_i||_2^2 + || x'-\mathrlap{m_j||_2^2)]}  \\ 
%     s.t. \quad Z_m = \int \exp[-\rho || x-m||_2^2] p(m) \; \textrm{d}m \; \approx \frac{1}{J}\sum_{i=1}^J \exp[-\rho || x-m_i||_2^2]   \numberthis \label{eq:approximation}
% \end{align*}
% \begin{align*} \numberthis \label{eq:approximation}
%     & k_{\textrm{WEG}}(x,x') \approx \frac{1}{Z_m Z_{m'} J^2} \quad \quad\quad \quad  \quad \quad \quad \quad \\
%     & \quad \quad \quad \quad  \times \sum_{i=1}^J \sum_{j=1}^J \exp[-\lambda d_{geo}(m_i, m_j)] \\
%     & \quad \quad \quad \quad \times \exp[-\rho (|| x-m_i||_2^2 + || x'-m_j||_2^2)]
% \end{align*}
% \begin{align*}\label{eq:approximation}
%     k_{\textrm{WEG}}(x,x') =  \mathbb{E}_{m,m' \sim  p(m)} \Big[\exp[-\lambda d_{geo}(m_i, m_j)] \nonumber \\
%     \times \exp[-\rho (|| x-m_i||_2^2 + || x'-m_j||_2^2)]\Big] \quad \quad \quad \numberthis
% \end{align*}
\begin{align*}\label{eq:approximation}
    k_{\textrm{WEG}}(x,x') \approx \frac{1}{Z_m Z_{m'} J^2}  \sum_{i=1}^J \sum_{j=1}^J \exp[-\lambda d_{geo}(m_i, m_j)] \nonumber \\
    \times \exp[-\rho (|| x-m_i||_2^2 + || x'-m_j||_2^2)] \quad \quad \quad \numberthis
\end{align*}
% \begin{equation}\label{eq:approximation}
%     k_{\textrm{WEG}}(x,x') \approx \frac{1}{Z_m Z_{m'} J^2}  \sum_{i=1}^J \sum_{j=1}^J \exp[-\lambda d_{geo}(m_i, m_j)] \nonumber
% \end{equation}
% \begin{equation}
%     \times \exp[-\rho (|| x-m_i||_2^2 + || x'-m_j||_2^2)] \quad \quad \quad \numberthis
% \end{equation}
% \begin{align*} 
%     \frac{1}{J^2} \sum_{i=1}^J \sum_{j=1}^J \underbrace{\exp[-\lambda d_{geo}(m_i, m_j)]}_{k_{\textrm{EG}}(m_i, m_j)} \underbrace{\frac{1}{Z_m }\exp[-\rho || x-m_i||_2^2]}_{\propto q(m|x,\rho)} \underbrace{\frac{1}{Z_{m'} }\exp[-\rho|| x'-m_j||_2^2]}_{\propto q(m|x,\rho)}  \\ 
%     s.t. \quad Z_m = \int \exp[-\rho || x-m||_2^2] p(m) \; \textrm{d}m \; \approx \frac{1}{J}\sum_{i=1}^J \exp[-\rho || x-m_i||_2^2]   \numberthis \label{eq:approximation}
% \end{align*}
We can similarly estimate constants $Z_m$, $Z_{m'}$:
% \begin{align*} \label{eqn:normalization}
%     Z_m = \int \exp[-\rho || x-m||_2^2] p(m) \; \textrm{d}m \\
%     \approx \frac{1}{J}\sum_{i=1}^J \exp[-\rho || x-m_i||_2^2] 
%     \numberthis
% \end{align*}
\begin{equation} \label{eqn:normalization}
    Z_m \approx \frac{1}{J}\sum_{i=1}^J \exp\left[-\rho || x-m_i||_2^2\right] 
\end{equation}
% \vspace{-5pt}
% \vspace{-2pt}
\subsection{GPEC Algorithm} \label{sec:alg}
\vspace{-1mm}
GPEC has separate training (Alg. \ref{alg:training}) and inference (Alg. \ref{alg:inference}) stages.
During training, GPEC constructs the EG kernel matrix by sampling the DB.
Note that in Eq. \eqref{eq:approximation} we calculate $k_{\textrm{EG}}(m_i, m_j)$ independently of $x$ and $x'$ $\: \forall i,j \in \{1...J\}$. Therefore, the EG kernel only needs to be calculated once for a set of DB samples. During training and inference, the WEG kernel weights the precalculated EG kernel based on $x,x'$ (Fig \ref{fig:kernel_comparison}C).
Once the GPEC model is trained, either the variance or confidence interval width of the predictive distribution can be used as the uncertainty estimate.
The training cost of GPEC is amortized during inference; GP inference generally has time complexity of $\mathcal{O}(N^3)$, which can be reduced to $\mathcal{O}(N^2)$ using BBMM \citep{gpytorch}, and further with variational methods (e.g., \citet{gal_variational_GP}).

DB sampling and geodesic distance estimation are ongoing areas of research. In our implementation, we adapt DeepDIG \citep{karimi2019characterizing} for sampling the DB of neural networks and DBPS \citep{DBPS_decision_boundary} for all other models. We utilize ISOMAP \citep{tenenbaumGlobalGeometricFramework2000} for estimating geodesic distances. Additional implementation detail is provided in App. \ref{app:implementation_details}.

\begin{figure*}[t]
   \begin{center}
   \includegraphics[width=1\linewidth]{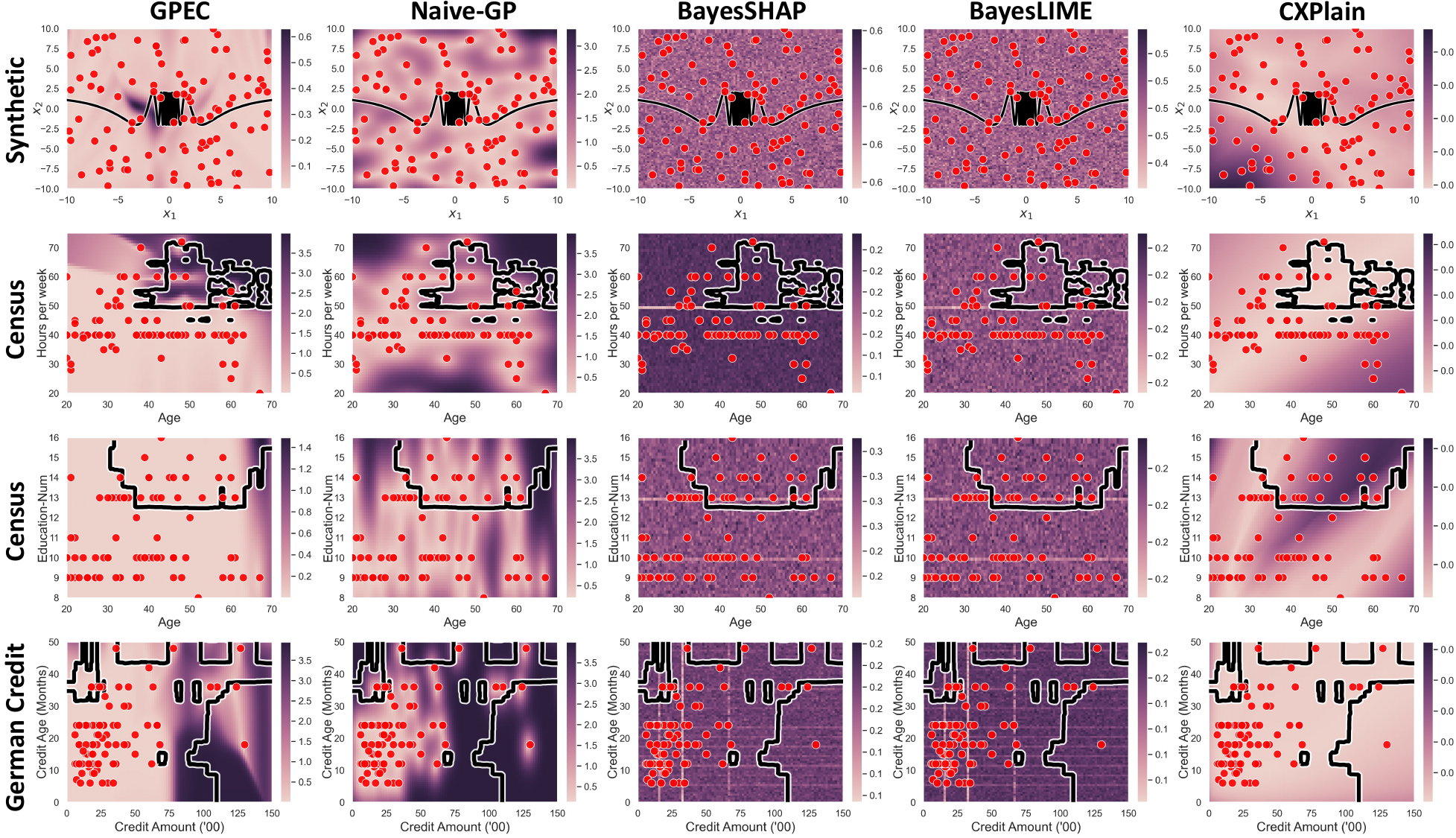}
   \end{center}
   \vspace{-3mm}
   \caption{Visualization of estimated explanation uncertainty for different datasets and competing methods. The heatmap represents uncertainty level for a grid of explanations for the x-axis feature; \emph{darker heatmap regions represent higher uncertainty}. The black line represents the black-box DB, and red points represent training samples. The heatmap shows that GPEC uncertainty is elevated for samples near complex decision boundaries. In contrast, heatmaps for BayesSHAP, BayesLIME, and CXPlain are relatively uniform.}
   \label{fig:uncertaintyfigure}
\end{figure*}

\vspace{-1mm}
\section{EXPERIMENTS}
\vspace{-2mm}
%---------------------------------------------
We evaluate GPEC on a variety of datasets and classifiers.
In section \ref{sec:experiment_uncertainty_visualization} we visually compare GPEC uncertainty with competing models. 
Section \ref{sec:experiment_regularizationtest} evaluates how GPEC captures DB complexity.
Section \ref{sec:experiment_fullfat} is an ablation test that disentangles the two sources of uncertainty.
All experiments were run on an internal cluster using AMD EPYC 7302 16-Core processors. CIFAR10 results were run on Nvidia A100 GPUs. All source code is available at \url{https://github.com/davinhill/GPEC}.

% \vspace{-5pt}
\vspace{-1mm}
\subsection{Experiment Setup}
\vspace{-2mm}

% \begin{algorithm}[tb]
\begin{algorithm}[t]
\small
\setlength{\textfloatsep}{10pt}
   \caption{GPEC Training}
   \label{alg:training}
    \textbf{Input :} Training Samples $X \in \mathbb{R}^{N \times D}$, Explainer. \\ 
    \textbf{Output :} WEG Kernel $K \in [0,1]^{N \times N}$, Explainer  Variance $U \in \mathbb{R}^{N \times D}_{+}$, Weighting $W\in [0,1]^{N \times J}$, EG Kernel $G \in [0,1]^{J \times J}$, DB Samples $M \in \mathbb{R}^{J \times D}$. \\

    Get Explainer Variance $U$ from Explainer \\
    Draw $J$ DB samples $M$ from DB Sampling Function \\
    \For{each pair of DB samples $m_i$, $m_j \in M$}{
        $G_{i,j} \leftarrow \exp(-\lambda  d_{geo}(m_i, m_j))$ $ \quad \backslash \backslash$ Eq. \eqref{eqn:eg}
    }
    %(Eq. \eqref{eqn:eg})
    \For{each data sample $x_i$ and DB sample $m_j$}{
        $W_{i,j} \leftarrow \exp(-\rho  || x_i - m_j ||_2^2)$ $\quad \backslash \backslash$ Eq. \eqref{eq:q_weighting}
    }
    % $W_{i,:} \leftarrow  \displaystyle \frac{W_{i,:}}{\sum_{j=1}^P W_{i,j}}$ $\quad \backslash \backslash$ Normalize weighting \\ 
    % $W_{i,:} \leftarrow  W_{i,:} / \sum_{j=1}^P W_{i,j}$ $\quad \backslash \backslash$ Normalize weighting \\ 
    $W_{i,:} \leftarrow \frac{W_{i,:}}{\sum_{j=1}^P W_{i,j}}$ $\quad \backslash \backslash$ Normalize weighting \\ 
    % $W_{i,:} \leftarrow  \displaystyle W_{i,:} \mathbin{/} \sum_{j=1}^P W_{i,j}$ $\quad \backslash \backslash$ Eq. \eqref{eqn:normalization} \\ 
    
    $K \leftarrow W G W^\intercal$  $\quad \quad \quad \backslash \backslash$ WEG Kernel \\
    Return $K$, $U$, $W$, $G$, $M$
\end{algorithm}
% \begin{algorithm}[tb]
\setlength{\textfloatsep}{10pt}

\begin{algorithm}[t]
\small
   \caption{GPEC Inference}
   \label{alg:inference}
    \textbf{Input :} Sample $x \in \mathbb{R}^D$; $K$, $U$, $W$, $G$, $M$ from Alg. \ref{alg:training}. \\ 
    \textbf{Output :} GPEC Uncertainty $V \in \mathbb{R}^{D}_{+}$ \\
    \\
    $\backslash \backslash$ Calculate weighting Eq. \eqref{eq:q_weighting} \\
    \For{each DB sample $m_i \in M$}{
        $W^*_{i} \leftarrow \exp(-\rho  || x - m_i ||_2^2)$
    }
    $W^* \leftarrow \frac{W^*}{\sum_{i=1}^P W_{i}}$ \\ 
    % $\backslash \backslash$ Eq. \eqref{eqn:GP_variance} \\
    \\
    \For{each explanation dimension $d \in D$}{
        $V_d = W^* G W^{* \intercal} \\ - W^*G W^{\intercal}[K + I_N U_{:,d}]^{-1} W^{\intercal} G W^{*}$   
    }
    Return $V$
    % \vspace{-5mm}
\end{algorithm}

Unless otherwise stated, we set $\lambda = 1.0$ and $\rho = 0.1$ (see App. \ref{app:sensitivity_analysis} for experiments on parameter sensitivity), and use GPEC with the KernelSHAP explainer.

\textbf{Datasets.} Experiments are performed on three tabular datasets (Census, Online Shoppers \citep{sakarRealtimePredictionOnline2019}, German Credit) from the UCI data repository \citep{uci_datasets}, and three image datasets (MNIST \citep{lecun-mnisthandwrittendigit-2010}, f-MNIST \citep{fmnist}), and CIFAR10 \citep{cifar10}).
We additionally create a toy example (Synthetic) where training and test samples are drawn from the uniform distribution over $[-10,10]^2$ and the model DB is defined as follows:
\begin{align*}
    \mathcal{M}_{\textrm{synth}} = \{\left(x_1, f(x_1)\right): x_1 \in \mathbb{R}\} \quad \quad \quad \quad \\
    f(x_1) = \begin{cases} 
      2\cos (\frac{10}{x_1}) & |x_1| \geq \frac{20}{(5\textrm{e}6+1)\pi} \\
      0 &  |x_1| < \frac{20}{(5\textrm{e}6+1)\pi}
   \end{cases}
\end{align*}
GPEC can be used with any black-box model; we use XGBoost \citep{xgboost} for tabular datasets, 4-layer neural network for MNIST and f-MNIST, and Resnet18 \citep{resnet} for CIFAR10. Additional dataset details are outlined in App. \ref{app:dataset_details}.

\textbf{Comparisons. }We compare GPEC to a baseline GP implementation plus three other competing explanation uncertainty estimation methods.
Naive-GP similarly uses a GP parametrization (Eq. \eqref{eqn:GP}) but instead uses the Radial Basis Function kernel, which does not incorporate DB information.
BayesSHAP and BayesLIME \citep{reliable:neurips21} are extensions of KernelSHAP and LIME, respectively, that fit local Bayesian linear regression models.
CXPlain \citep{cxplain} trains a surrogate model and uses bootstrapping to estimate explanation uncertainty.
Additional details on competing methods are outlined in App. \ref{app:competitor_details}.

% \vspace{-5pt}
\vspace{-5pt}
\subsection{Uncertainty Visualization} \label{sec:experiment_uncertainty_visualization}
\vspace{-5pt}

\begin{table*}[t]
    %------------------------------------------
    % \setlength{\belowcaptionskip}{-15pt}

    \scriptsize
    % \small
    \setlength{\tabcolsep}{2.3pt}
    \renewcommand{\arraystretch}{1.1}
    \centering
    \begin{tabular}{ l c c c c c c c c c c c c c c c}
        %--------------
        \cline{0-15}
        \multicolumn{1}{|l|}{\emph{Dataset}} & 
        \multicolumn{3}{c|}{\textbf{Census}}  & 
        \multicolumn{3}{c|}{\textbf{Online Shoppers}} &
        \multicolumn{3}{c|}{\textbf{German Credit}} &
        \multicolumn{6}{c|}{\textbf{CIFAR10}} \\
        % \multicolumn{3}{c}{\textbf{ }} \\
        \cline{0-15}

        %--------------
        \multicolumn{1}{|l|}{\emph{Regularization}} & 
        \multicolumn{3}{c|}{$\gamma$}  & 
        \multicolumn{3}{c|}{$\gamma$} &
        \multicolumn{3}{c|}{$\gamma$} &
        \multicolumn{3}{c|}{$\ell_2$} &
        \multicolumn{3}{c|}{Softplus $\beta$} \\
        \cline{0-15}
        %--------------
        \multicolumn{1}{|l|}{\emph{Magnitude}} & 
        \multicolumn{1}{c}{$0$}  & 
        \multicolumn{1}{c}{$5$}  & 
        \multicolumn{1}{c|}{$10$}  & 
        \multicolumn{1}{c}{$0$}  & 
        \multicolumn{1}{c}{$5$}  & 
        \multicolumn{1}{c|}{$10$}  & 
        \multicolumn{1}{c}{$0$}  & 
        \multicolumn{1}{c}{$5$}  & 
        \multicolumn{1}{c|}{$10$} &
        \multicolumn{1}{c}{$0$}  & 
        \multicolumn{1}{c}{$1\textrm{e-5}$}  & 
        \multicolumn{1}{c|}{$10\textrm{e-5}$}  & 
        \multicolumn{1}{c}{$1.0$}  & 
        \multicolumn{1}{c}{$0.5$}  & 
        \multicolumn{1}{c|}{$0.25$} \\
        \cline{0-15} 
        %--------------
            
        \multicolumn{1}{|l|}{GPEC} &
            \bcolor{1.573} &
            \bcolor{1.177} &
            \multicolumn{1}{c|}{\bcolor{1.158}}  &
            %---
            \bcolor{0.209} &
            \bcolor{0.123} &
            \multicolumn{1}{c|}{\bcolor{0.092}}  &
            %---
            \bcolor{2.665} &
            \bcolor{1.747} &
            \multicolumn{1}{c|}{\bcolor{0.250}}  & 
            %---
            \bcolor{0.029} &
            \bcolor{0.029} &
            \multicolumn{1}{c|}{\bcolor{0.028}}  & 
            %---
            \bcolor{0.033} &
            \bcolor{0.032} &
            \multicolumn{1}{c|}{\bcolor{0.032}}   

            \\

        %--------------
        \multicolumn{1}{|l|}{Naive-GP} &
            \textbf{0.498} &
            \textbf{0.472} &
            \multicolumn{1}{c|}{\textbf{0.467}}  &
            %---
            3.699 &
            3.724 &
            \multicolumn{1}{c|}{3.730}  &
            %---
            0.330 &
            0.412 &
            \multicolumn{1}{c|}{2.533} & 
            %---
            0.902 &
            0.902 &
            \multicolumn{1}{c|}{0.902} & 
            %---
            0.903 &
            0.903 &
            \multicolumn{1}{c|}{0.903}  
            \\

        %--------------
        \multicolumn{1}{|l|}{BayesSHAP} &
            0.037 &
            0.037 &
            \multicolumn{1}{c|}{0.037}  &
            %---
            0.031 &
            0.031 &
            \multicolumn{1}{c|}{0.031}  &
            %---
            0.019 &
            0.019 &
            \multicolumn{1}{c|}{0.018}  & 
            %---
            -- &
            -- &
            \multicolumn{1}{c|}{--}  & 
            %---
            -- &
            -- &
            \multicolumn{1}{c|}{--}   
            \\
            
        %--------------
        \multicolumn{1}{|l|}{BayesLIME} &
            \textbf{0.097} &
            \textbf{0.096} &
            \multicolumn{1}{c|}{\textbf{0.095}}  &
            %---
            \textbf{0.098} &
            \textbf{0.097} &
            \multicolumn{1}{c|}{\textbf{0.093}}  &
            %---
            \textbf{0.085} &
            \textbf{0.066} &
            \multicolumn{1}{c|}{\textbf{0.045}}  & 
            %---
            -- &
            -- &
            \multicolumn{1}{c|}{--}  & 
            %---
            -- &
            -- &
            \multicolumn{1}{c|}{--}   
            \\
            
        %--------------
        \multicolumn{1}{|l|}{CXPlain} &
            0.064 &
            0.064 &
            \multicolumn{1}{c|}{0.069}  &
            %---
            0.004 &
            0.003 &
            \multicolumn{1}{c|}{0.006} &
            %---
            1.7$\textrm{e-4}$ &
            1.1$\textrm{e-4}$ &
            \multicolumn{1}{c|}{4.3$\textrm{e-4}$}  & 
            %---
            9.5$\textrm{e-5}$ &
            0.2$\textrm{e-5}$ &
            \multicolumn{1}{c|}{2.6$\textrm{e-5}$}  & 
            %---
            1.6$\textrm{e-5}$ &
            0.2$\textrm{e-5}$ &
            \multicolumn{1}{c|}{9.1$\textrm{e-5}$} 
            \\

        %--------------
        \cline{0-15}
        &
            &
            &
            &
            %---
            &
            &
            &
            %---
            &
            & 
            & 
            %---
            &
            & 
            & 
            %---
            &
            & 
            \\
        %--------------

        \cline{0-12}

        \multicolumn{1}{|l|}{\emph{Dataset}} & 
        \multicolumn{6}{c|}{\textbf{MNIST}}  & 
        \multicolumn{6}{c|}{\textbf{Fashion MNIST}}
         \\
        \cline{0-12}
        %--------------
        \multicolumn{1}{|l|}{\emph{Regularization}} & 
        \multicolumn{3}{c|}{$\ell_2$}  & 
        \multicolumn{3}{c|}{Softplus $\beta$}  & 
        \multicolumn{3}{c|}{$\ell_2$}  & 
        \multicolumn{3}{c|}{Softplus $\beta$} 
         \\
        \cline{0-12}
        %--------------
        \multicolumn{1}{|l|}{\emph{Magnitude}} & 
        \multicolumn{1}{c}{$0$}  & 
        \multicolumn{1}{c}{$1\textrm{e-5}$}  & 
        \multicolumn{1}{c|}{$10\textrm{e-5}$}  & 
        \multicolumn{1}{c}{$1.0$}  & 
        \multicolumn{1}{c}{$0.5$}  & 
        \multicolumn{1}{c|}{$0.25$}  & 
        \multicolumn{1}{c}{$0$}  & 
        \multicolumn{1}{c}{$1\textrm{e-5}$}  & 
        \multicolumn{1}{c|}{$10\textrm{e-5}$}  & 
        \multicolumn{1}{c}{$1.0$}  & 
        \multicolumn{1}{c}{$0.5$}  & 
        \multicolumn{1}{c|}{$0.25$} 
         \\
        \cline{0-12}
        %--------------

        \multicolumn{1}{|l|}{GPEC} &
            \bf{0.236} &
            \bf{0.157} &
            \multicolumn{1}{c|}{\bf{0.078}}  &
            %---
            \bcolor{0.087} &
            \bcolor{0.073} &
            \multicolumn{1}{c|}{\bcolor{0.056}}  &
            %---
            \bcolor{0.378} &
            \bcolor{0.187} &
            \multicolumn{1}{c|}{\bcolor{0.063}}  &
            %---
            \bcolor{0.112} &
            \bcolor{0.075} &
            \multicolumn{1}{c|}{\bcolor{0.061}} 

            \\

        %--------------
        \multicolumn{1}{|l|}{Naive-GP} &
            4.00 &
            4.00 &
            \multicolumn{1}{c|}{3.99}  &
            %---
            0.226 &
            0.232 &
            \multicolumn{1}{c|}{0.236}  &
            %---
            3.98 &
            3.99 &
            \multicolumn{1}{c|}{3.99}  &
            %---
            0.301 &
            0.261 &
            \multicolumn{1}{c|}{0.262} 

            \\

        %--------------
        \multicolumn{1}{|l|}{BayesSHAP} &
            \textbf{0.025} &
            \textbf{0.016} &
            \multicolumn{1}{c|}{\textbf{0.008}}  &
            %---
            \textbf{0.013} &
            \textbf{0.011} &
            \multicolumn{1}{c|}{\textbf{0.010}}  &
            %---
            \textbf{0.030} &
            \textbf{0.014} &
            \multicolumn{1}{c|}{\textbf{0.007}}  &
            %---
            \textbf{0.018} &
            \textbf{0.010} &
            \multicolumn{1}{c|}{\textbf{0.009}} 

            \\
            
        %--------------
        \multicolumn{1}{|l|}{BayesLIME} &
            \bcolor{2.452} &
            \bcolor{1.573} &
            \multicolumn{1}{c|}{\bcolor{0.737}}  &
            %---
            0.868 &
            0.866 &
            \multicolumn{1}{c|}{0.721}  &
            %---
            \textbf{2.605} &
            \textbf{1.364} &
            \multicolumn{1}{c|}{\textbf{0.666}}  &
            %---
            \textbf{1.178} &
            \textbf{0.861} &
            \multicolumn{1}{c|}{\textbf{0.779}} 

            \\
            
        %--------------
        \multicolumn{1}{|l|}{CXPlain} &
            0.1$\textrm{e-5}$ &
            5.0$\textrm{e-5}$ &
            \multicolumn{1}{c|}{8.6 $\textrm{e-5}$}  &
            %---
            5.3$\textrm{e-5}$ &
            8.0$\textrm{e-5}$ &
            \multicolumn{1}{c|}{5.4$\textrm{e-5}$}  &
            %---
            7.2$\textrm{e-5}$ &
            4.8$\textrm{e-5}$ &
            \multicolumn{1}{c|}{6.2 $\textrm{e-5}$}  &
            %---
            9.0$\textrm{e-5}$ &
            6.6$\textrm{e-5}$ &
            \multicolumn{1}{c|}{9.6$\textrm{e-5}$} 

            \\
        %--------------
    
        \cline{0-12}
    \end{tabular}
    \caption{Average explanation uncertainty for classifiers with increasing (left to right) levels of regularization, which controls relative model complexity. We evaluate how well GPEC reflects model complexity; increased regularization should result in lower uncertainty. Methods that have decreasing estimates are bolded; the method with the greatest percentage decrease is highlighted in blue. CIFAR10 results for BayesSHAP and BayesLIME are omitted due to computational expense.}
    \label{table:regularize}
    
    \end{table*}
    
To visualize explanation uncertainty, we plot uncertainty estimates as a heatmap for the explanations derived from an XGBoost binary classifier trained on two selected features. 
Figure \ref{fig:uncertaintyfigure} plots the uncertainty heatmap for the x-axis feature (y-axis feature results shown in App. \ref{app:addnl_results_uncertainty_visualization}), where \emph{darker heatmap regions indicate higher uncertainty}.
Red points represent training samples for GPEC, Naive-GP, and CXPlain, and represent background samples for BayesSHAP and BayesLIME. The DB is plotted as a black line.

\begin{figure}[t]
    \begin{center}
    \includegraphics[width=1\linewidth]{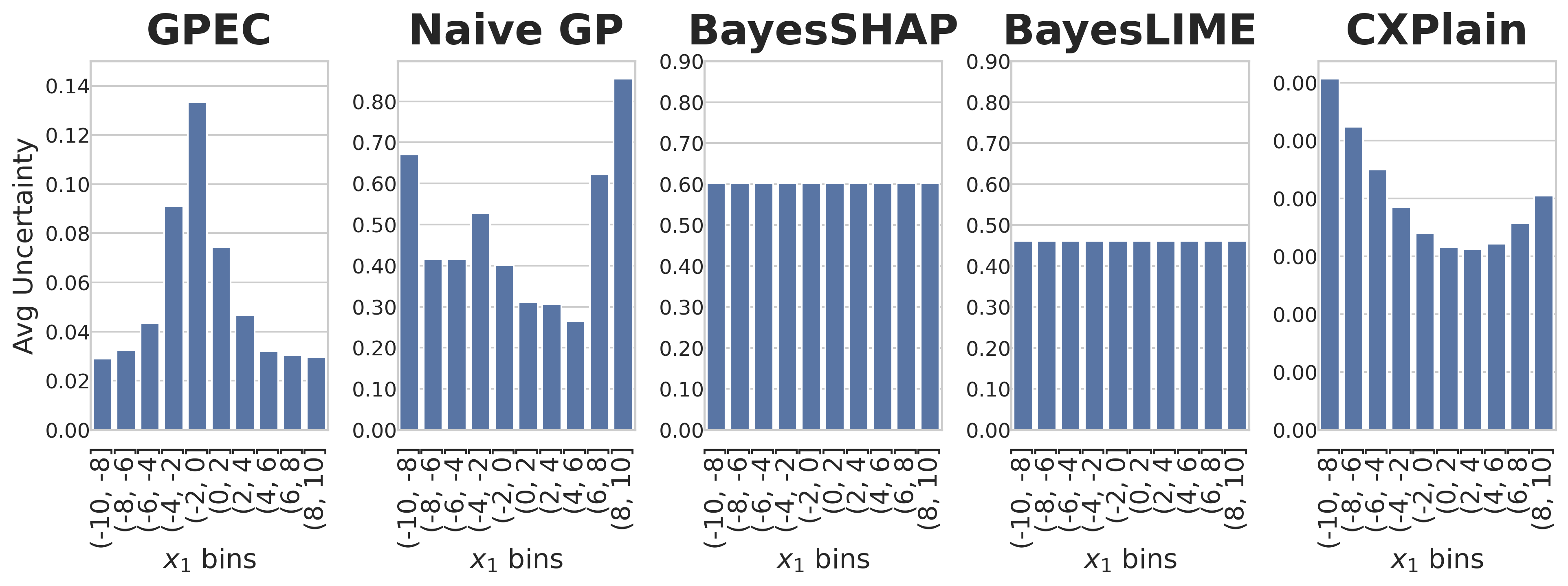}
    \end{center}
    \vspace{-4mm}
    \caption{Average uncertainty values for different regions in Synthetic, binned by $x_1$. Synthetic is designed to have higher DB complexity for $x_1 \in [-4,4]$, which is reflected by high GPEC uncertainty in bins $(-4,2]$, $(-2,0]$, $(0,2]$, $(2,4]$. Other methods do not capture the high DB complexity for $x_1 \in [-4,4]$.}
    \label{fig:cosinv}
    % \vspace{-2mm}
\end{figure}

We expect to see higher GPEC uncertainty (dark heatmap regions) for test samples farther away from training samples (red) and close to nonlinearities in the DB.
We observe that this holds true, especially for high uncertainty regions in the center of Synthetic (top row), the top-right of of Census (2$^{\textrm{nd}}$ and 3$^{\textrm{rd}}$ rows), and the bottom-right of German Credit (bottom).
% Comparing GPEC and Naive-GP shows that using the WEG kernel attributes higher uncertainty to samples near nonlinearities in the DB.
In particular, Synthetic is constructed such that the DB for $x_1 \in [-4,4]$ is more complex than $x_1 \notin [-4,4]$.
To highlight this, in Figure \ref{fig:cosinv} we bin the values of $x_1$ and calculate the average uncertainty over each bin.
Indeed we observe that the bins within $[-4,4]$ exhibit the highest average uncertainty values under GPEC.

In contrast to GPEC, Naive-GP provides uncertainty estimates that relate only to the training samples; test sample uncertainty is proportional to distance from the training samples.
The competing methods BayesSHAP, BayesLIME, and CXPlain result in relatively uniform uncertainty estimates over the test samples. CXPlain shows areas of higher uncertainty for Census, however the magnitude of these estimates are small.
The uncertainty estimates produced by these competing methods are unable to capture the properties of the black-box model.
%---------------------------------------------
% \vspace{-5pt}
\vspace{-8pt}
\subsection{Regularization Test} \label{sec:experiment_regularizationtest}
\vspace{-6pt}

In this section we evaluate how well GPEC captures uncertainty due to DB complexity.
DB and model complexity is generally difficult to quantify; we instead use regularization methods to control relative model complexity. By examining the average uncertainty across different models, we can better understand how well GPEC uncertainty reflects the underlying DB complexity.
For XGBoost models, we vary $\gamma$, which penalizes the number of leaves in each tree \citep{xgboost}.
For neural networks, we regularize: 1) $\ell_2$ penalty on the weights, and 2) we change the ReLU activation to Softplus: a smooth approximation of ReLU with smoothness inversely proportional to parameter $\beta$ \citep{dombrowski2019explanations}.

Results are presented in Table \ref{table:regularize} (standard error and parameters reported in App. \ref{app:addnl_results_regularization}).
GPEC shows a decreasing average uncertainty with increased regularization, indicating its ability to reflect the complexity of the underlying black-box model.
For tabular datasets, the estimates for BayesSHAP, BayesLIME, and CXPlain stay relatively flat.
Interestingly, the estimates from these methods decrease for the image datasets; we hypothesize that the neural network regularization also increases overall stability of the explanations.
% GPEC can capture both the uncertainty from WEG kernel and also the estimated uncertainty from the function approximation for the explainer, which we demonstrate in Section \ref{sec:experiment_fullfat}. \todo{this sentence is awkward}
%---------------------------------------------
% \vspace{-5pt}
\vspace{-8pt}
\subsection{GPEC Ablation Test} \label{sec:experiment_fullfat}
\vspace{-6pt}

\begin{figure*}
  \begin{minipage}[c]{0.6\textwidth}
  \includegraphics[width=1\linewidth]{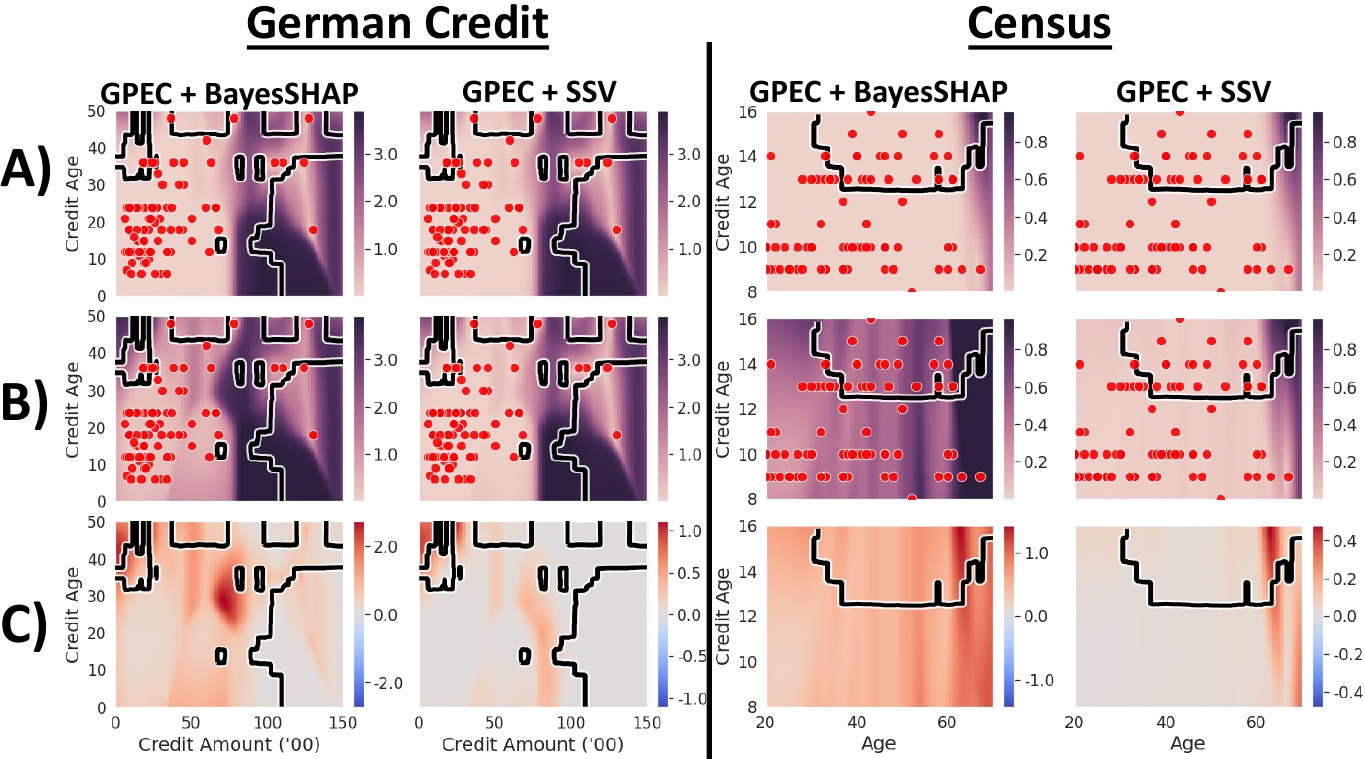}
  \end{minipage}\hfill
  \begin{minipage}[c]{0.37\textwidth}
 \caption{Ablation test. \textbf{Row (A)} visualizes the GPEC uncertainty estimate using BayesSHAP and SSV. \textbf{Row (B)} shows the DB-aware uncertainty component in GPEC for the estimate in row (A). \textbf{Row (C)} subtracts row (B) from row (A), which isolates the function approximation uncertainty in GPEC. GPEC is able to combine and also disentangle the two sources of uncertainty.}
     \label{fig:fullfat}
  \end{minipage}
  % \vspace{-3mm}
\end{figure*}

GPEC can capture both the uncertainty from WEG kernel and also the estimated uncertainty from the noisy explanation labels.
These noisy explanation labels can either originate directly from the explainer, or can be estimated empirically (Eq. \eqref{eqn:empirical_uncertainty}).
Here, we calculate GPEC uncertainty with two different explainers, BayesSHAP and Shapley Sampling Values (SSV) \citep{shapley_sampling}, and ablate the DB-aware uncertainty in order to evaluate function approximation uncertainty. The two selected explainers are different SHAP approximations; the former has an in-built uncertainty estimate whereas we use empirical sampling (Eq. \eqref{eqn:empirical_uncertainty}) for SSV. In Figure \ref{fig:fullfat} row (A) we plot the heatmap for the combined GPEC estimate. In row (B) we exclusively show the DB-aware uncertainty by training GPEC with noiseless explanations. The difference (i.e. the effects of function approximation uncertainty) is shown in the row (C). We observe that the combined GPEC estimate is able to effectively integrate and disentangle both sources of uncertainty. Interestingly, the BayesSHAP explanations have higher function approximation uncertainty, which results in higher GPEC estimates. This suggests that users wanting to reduce explanation uncertainty might prefer SSV over BayesSHAP in this scenario.

%=================================
\vspace{-2mm}
\subsection{Time Complexity} \label{sec:time}
\vspace{-1mm}
In Table \ref{table:time} we show an execution time comparison for generating uncertainty estimates for 100 explanations (inference time). During inference, GPEC is comparable to methods that amortize training time (Naive-GP and CXPlain) and is significantly faster than perturbation methods (BayesSHAP and BayesLIME).
Execution time results for training are shown in App. \ref{app:execution_time}.
\begin{table}[t]
        %------------------------------------------
        % \setlength{\belowcaptionskip}{-15pt}
    
        % \setlength{\abovecaptionskip}{5pt}
        % \small
        \scriptsize
        \setlength{\tabcolsep}{2.0pt}
        \renewcommand{\arraystretch}{1.1}
        \centering
        \begin{tabular}{ l  c  c  c  c   c c }
            %--------------
            % \toprule
            \hline
            & 
            \multicolumn{1}{c}{\textbf{Census}}  & 
            \multicolumn{1}{c}{\textbf{Shop}} &
            \multicolumn{1}{c}{\textbf{Credit}} &
            % \multicolumn{1}{C{1cm}}{Online Shoppers} &
            % \multicolumn{1}{C{1cm}}{German Credit} &
            \multicolumn{1}{c}{\textbf{MNIST}} &
            \multicolumn{1}{c}{\textbf{fMNIST}} &
            \multicolumn{1}{c}{\textbf{CIFAR}}
            % \multicolumn{1}{c}{ImNet}
            \\
            \hline 
    
            %--------------
                
            GPEC &
                0.11 &
                0.37 &
                0.07 &
                12.90 &
                18.15 &
                1.77
                % 1.77
    
                \\
    
            %--------------
            Naive-GP &
                0.00 &
                0.00 &
                0.02 &
                8.95 &
                7.41 &
                1.29
                % 1.29 
    
                \\
    
            %--------------
            CXPlain &
                0.05 &
                0.06 &
                0.04 &
                9.76 &
                18.18 &
                8.27
                % 8.27
    
                \\
            %--------------
            BayesSHAP &
                140.40 &
                54.56 &
                4.86 &
                42,467 &
                42,361 &
                --
                % --
    
                \\
    
            %--------------
            BayesLIME &
                91.29 &
                54.60 &
                4.83 &
                41,832 &
                41,992 &
                --
                % --
                \\
        
            \hline
        \end{tabular}
       \vspace{-1mm}
        \caption{Execution time comparison (inference) for uncertainty estimation of 100 samples, in seconds. CIFAR10 results for BayesLIME and BayesSHAP are ommitted due to computational cost. CIFAR10 results use a single Nvidia A100 GPU; all other results use CPU only. }
        \label{table:time}
        \vspace{3mm}
\end{table} 

%=================================
\vspace{-2mm}
\subsection{Additional Results} \label{sec:addl_results}
\vspace{-1mm}
Due to space constraints, we include additional experiments in the appendix, including a case study on diabetes prediction (App. \ref{app:diabetes}), illustrative examples for image datasets (App. \ref{app:illustrative_examples}), and parameter sensitivity analysis (App. \ref{app:sensitivity_analysis}).

% \vspace{-6pt}
% \section{LIMITATIONS AND CONCLUSION}
\section{CONCLUSION}
\vspace{-2mm}
% \vspace{-8pt}
Generating uncertainty estimates for feature attribution explainers is essential for establishing reliable explanations.
We introduce a novel GP-based approach that can be used with any black-box classifier and feature attribution method.
GPEC generates explanation uncertainty that combines 1) boundary-aware uncertainty, which captures the complexity of the DB, and 2) functional approximation uncertainty.
Experiments show that capturing this uncertainty improves understanding of the explanations and the black-box model itself.

Regarding limitations, GPEC relies on DB estimation methods which is an ongoing area of research. Due to the time complexity of DB estimation, this can result in a tradeoff between computation time and approximation accuracy or sample bias.
However, the effects of DB sampling time are minimized during inference as the DB only needs to be sampled during training.
Additionally, in its current implementation GPEC is limited to classification; we leave the extension to regression as future work.
\subsubsection*{Acknowledgements}
% This work was supported in part by NIH 2T32HL007427-41 from the National Heart, Lung, and Blood Institute, and by the Institute of Experiential AI at Northeastern University.
This work was supported in part by NIH 2T32HL007427-41 and U01HL089856 from the National Heart, Lung, and Blood Institute; NIH R01CA240771 and U24CA264369 from the National Cancer Institute; and the Institute of Experiential AI at Northeastern University.

% COPD Grant

% \clearpage
\bibliography{GP_Explanation_Uncertainty,manual_bibtex}
\bibliographystyle{unsrtnat}

%%%%%%%%%%%%%%%%%%%%%%%%%%%%%%%%%%%%%%%%%%%%%%%%%%%%%%%%%%%%
\clearpage
\section*{Checklist}

% %%% BEGIN INSTRUCTIONS %%%
 \begin{enumerate}

 \item For all models and algorithms presented, check if you include:
 \begin{enumerate}
   \item A clear description of the mathematical setting, assumptions, algorithm, and/or model. [Yes]
   \item An analysis of the properties and complexity (time, space, sample size) of any algorithm. [Yes]
   \item (Optional) Anonymized source code, with specification of all dependencies, including external libraries. [Yes]
 \end{enumerate}

 \item For any theoretical claim, check if you include:
 \begin{enumerate}
   \item Statements of the full set of assumptions of all theoretical results. [Yes]
   \item Complete proofs of all theoretical results. [Yes]
   \item Clear explanations of any assumptions. [Yes]     
 \end{enumerate}

 \item For all figures and tables that present empirical results, check if you include:
 \begin{enumerate}
   \item The code, data, and instructions needed to reproduce the main experimental results (either in the supplemental material or as a URL). [Yes]
   \item All the training details (e.g., data splits, hyperparameters, how they were chosen). [Yes]
         \item A clear definition of the specific measure or statistics and error bars (e.g., with respect to the random seed after running experiments multiple times). [Yes]
         \item A description of the computing infrastructure used. (e.g., type of GPUs, internal cluster, or cloud provider). [Yes]
 \end{enumerate}

 \item If you are using existing assets (e.g., code, data, models) or curating/releasing new assets, check if you include:
 \begin{enumerate}
   \item Citations of the creator If your work uses existing assets. [Yes]
   \item The license information of the assets, if applicable. [Yes]
   \item New assets either in the supplemental material or as a URL, if applicable. [Not Applicable]
   \item Information about consent from data providers/curators. [Not Applicable]
   \item Discussion of sensible content if applicable, e.g., personally identifiable information or offensive content. [Not Applicable]
 \end{enumerate}

 \item If you used crowdsourcing or conducted research with human subjects, check if you include:
 \begin{enumerate}
   \item The full text of instructions given to participants and screenshots. [Not Applicable]
   \item Descriptions of potential participant risks, with links to Institutional Review Board (IRB) approvals if applicable. [Not Applicable]
   \item The estimated hourly wage paid to participants and the total amount spent on participant compensation. [Not Applicable]
 \end{enumerate}

 \end{enumerate}

\clearpage
\appendix
\onecolumn

\section{BROADER IMPACT STATEMENT}
Machine learning models are increasingly relied upon in a diverse set of high-impact domains, ranging from healthcare to financial lending \citep{esteva2019guide, kose2021segmentation, doshi2017towards, 9155614, ashutosh, deep_learning_utilizing_suboptimal_spirometry, bazzaz2023active}. Therefore, it is crucial that users of these models can accurately interpret why predictions are made. 
For instance, a doctor may wish to know if a skin-cancer classifier's high performance is due to the use of truly diagnostic features, or rather due to spurious imaging device artifacts. 
However, further spurred by the advent of deep learning's increasing popularity, many of the models being deployed in these high-stakes fields are opaque and complex black boxes, producing predictions which are non-trivial to understand.
Many methods for explaining black-box predictions have been developed to address this issue \citep{ribeiroLIME, shap_lundberg, covertFeatureRemovalUnifying2020, masoomi2022explanations, torop2023smoothhess}, but explanations may have varying quality and consistency. Before utilizing explanations in practice, it is essential that users know when, and when not, to trust them. Explanation uncertainty is one proxy for this notion of trust, in which more uncertain explanations may be deemed less trustworthy. In this work, we explore a new way to model explanation uncertainty, in terms of local decision-boundary complexity.  
In tandem with the careful consideration of domain experts, our methodology may be used to assist in determining when explanations are reliable. 

\section{BACKGROUND} \label{app:background}

\subsection{Related Works: Reliability of Explanations} \label{app:relatedworks_explanation_reliability}

% \textbf{Reliability of Explanations.}
While feature attribution methods have gained wide popularity, a number of issues relating to the reliability of such methods have been uncovered. \citet{alvarez-melis2018} investigate the notion of robustness and show that many feature attribution methods are sensitive to small changes in input. This has been further investigated in the adversarial setting for perturbation-based methods \citep{slack_foolingLIMESHAP} and neural network-based methods \citep{ghorbaniInterpretationNeuralNetworks2019}. \citet{Kindermans2019} show that many feature attribution methods are affected by distribution transformations such as those common in preprocessing. The generated explanations can also be very sensitive to hyperparameter choice \citep{bansalSAMSensitivityAttribution2020}.
A number of metrics have been proposed for evaluating explainer reliability, such as with respect to adversarial attack \citep{dombrowski2019explanations, ghorbaniInterpretationNeuralNetworks2019, hsiehEvaluationsMethodsExplanation2021}, local perturbations \citep{alvarez-melis2018, VSI_visani2022}, black-box smoothness \citep{zulqarnain_lipschitzness}, fidelity to the black-box model \citep{yehFidelitySensitivityExplanations2019}, or combinations of these metrics \citep{bhattEvaluatingAggregatingFeaturebased2020}.

\subsection{Gaussian Process Review} \label{app:GPReview}
A single-output Gaussian Process represents a distribution over \textit{functions} $f : \mathcal{X} \rightarrow \mathbb{R}$
\begin{equation}
f(x) \sim \mathcal{GP}(m(x), k(x,x')).
\end{equation}
where $m : \mathcal{X} \rightarrow \mathbb{R} $ and $k : (\mathcal{X}, \mathcal{X}) \rightarrow \mathbb{R}$ are the mean and kernel (or covariance) functions respectively, which are chosen \textit{a priori} to encode the users assumptions about the data. The kernel function $k(x,x')$ reflects a notion of similarity between data points for which predictive distributions over $f(x),f(x')$ respect.

The prior $m(x)$ -- frequently considered to be less important -- is commonly chosen to be the constant $m(x) = 0$.  

Specifically, a GP is an infinite collection of random variables $f(x)$, each indexed by an element $x \in \mathcal{X}$. Importantly, any finite sub-collection of these random variables 
\begin{equation}
f(X_{tr}) = (f(x_1) \ldots, f(x_n)) \in \mathbb{R}^D,
\end{equation}
corresponding to some index set $X_{tr} = \{x_i\}_{i=1}^n \subset \mathcal{X}$,  follows the multivariate normal distribution, i.e.
\begin{equation}
    f(X_{tr}) \sim \mathcal{N}(m(X_{tr}),K(X_{tr}, X_{tr})).
\end{equation}
The mean vector $m(X_{tr}) = (m(x_1), \ldots, m(x_n)) \in \mathbb{R}^n$ represents the mean function applied on each $x \in X_{tr}$ and the covariance matrix $K \in \mathbb{R}^{n\times n}$, also known as the gram matrix, contains each pairwise kernel-based similarity value $K_{ij} = k(x_i, x_j)$. Kernel function outputs correspond to dot products in potentially infinite dimensional expanded feature space, which allows for the encoding of nuanced notions of similarity; e.g. the exponential geodesic kernel referenced in this work \citep{feragen_geodesic_exponential_kernel}. 

Making predictions with a GP is analogous to simply conditioning this normal distribution on our data. Considering a set of input,noise-free label pairs
\begin{equation}
\mathcal{D} = \{ (x_i, f(x_i)\}_{i=1}^n    
\end{equation}
we may update our posterior over \textit{any subset} of the random variables $f(x)$ by considering the joint normal over the subset and $\mathcal{D}$ and conditioning on $\mathcal{D}$. For instance, when choosing a singleton index set $\{x_0\}$, the posterior over $f(x_0) | D$ is another normal distribution which may be written 
%for an arbitrary index point $x_0 \in \mathcal{X}$ 
as\footnote{assuming prior $m(x) =0$}
\begin{equation}
    f(x_0) \sim \mathcal{N}(\bar{f}(x_0), \mathbb{V}[f(x_0)]) 
\end{equation}
where
\begin{align}
    \bar{f}(x_0) &= K(x_0,X_{tr})^T K(X_{tr},X_{tr})^{-1} f(X_{tr})\\
    \mathbb{V}[f(x_0)] &= k(x_0,x_0) - K(x_0,X_{tr})^TK(X_{tr},X_{tr})^{-1}K(x_0,X_{tr}) \label{noise-free-var}
\end{align}
and $K(x_0,X_{tr}) \in \mathbb{R}^{D}$ is defined element-wise by $K(x_0,X_{tr})_i = k(x_0, x_i)$. 

Now we may consider the situation where our labels are noisy: 
\begin{equation}
\mathcal{D} = \{(x_i,y_i)\}_{i=1}^n , \  y_i = f(x_i) + \epsilon, \ \epsilon \sim \mathcal{N}(0,\sigma^2), \ \sigma^2 \in \mathbb{R}_+.
\end{equation}
Here $y_i$ is equal to the function output $f(x_i)$, with the addition of noise variable $\epsilon$. The conditional still follows a multivariate normal distribution, but the mean and variance equations are updated:
\begin{align}
    \bar{f}(x_0) &= K(x_0,X_{tr})^T (K(X_{tr},X_{tr}) + \sigma^2I)^{-1} Y\\
    \mathbb{V}[f(x_0)] &= k(x_0,x_0) - K(x_0,X_{tr})^T(K(X_{tr},X_{tr}) + \sigma^2I)^{-1}K(x_0,X_{tr}) \label{noise-var}
\end{align}
where $Y \in \mathbb{R}^{n}$ has elements $Y_i = y_i$. 

Note that the variance $\sigma^2 I$ is added to $K(X_{tr},X_{tr})$ in the quadratic form in Eq. \eqref{noise-var}, resulting in smaller eigenvalues after matrix inversion. Since this quadratic form is subtracted, the decision to model labels as noisy increases the uncertainty (variance) of the estimates that the GP posterior provides. This agrees with the intuition that noisy labels should result in more uncertain predictions. 
 
While GPs may also be defined over vector-valued functions, in this work the independence of each output component is assumed, allowing for modeling with $c\geq 1$ independent GPs. For more details see Ch.2 of \citet{GPML}.

\section{PROOF OF THEOREMS AND EXTENSIONS} \label{app:proof}

%%%%%%%%%%%%%%%%%%%%%%%%%%%%%%%%%%%%%%%%%%%%%%%%%
\subsection{Theorem \ref{thm:kernel_expgeodesic}: Relation to Exponential Geodesic Kernel} \label{proof:points_on_manifold}

\begin{align*}
    k(x,y) =  \int \int \exp[-\lambda d_{\textrm{geo}}(m, m')] q(m | x, \rho) q(m' | y, \rho)  \; \; \textrm{d} m' \textrm{d} m \\ 
    s.t. \quad q(m | x, \rho) \propto  \exp[-\rho || x-m||_2^2] p(m) \\
\end{align*}

Note that $\rho$ controls how to weight manifold samples close to $x, y$.
We take $\lim_{\rho \rightarrow \infty}$:

\[ \lim_{\rho \rightarrow \infty}  q(m | x, \rho)q(m' | y, \rho)= 
\begin{cases} 
    1 & x = m \; \textrm{and}\;  y = m' \\
    0 & \textrm{Otherwise}
 \end{cases}
\]

\iffalse
= \[ \begin{cases} 
    \int_{m \in \mathcal{M}_F} \int_{m' \in \mathcal{M}_F} \exp[-\lambda d_{\textrm{geo}}(m, m')] \; \; \textrm{d} m' \textrm{d}m & x = m \; \textrm{and}\;  y = m' \\
    0 & \textrm{Otherwise}
 \end{cases}
\]
\fi

Therefore the function within the integral of $k(x,y)$ evaluates to zero at all points except $x = m$ and $y=m'$. Since $x,y \in \mathcal{M}_F$ we can evaluate the integral:
$$k(x,y) = \exp[-\lambda d_{\textrm{geo}}(x, y)] $$

%%%%%%%%%%%%%%%%%%%%%%%%%%%%%%%%%%%%%%%%%%%%%%%%%
\subsection{Theorem \ref{thm:kernel_complexity}: Kernel Similarity and Decision Boundary Complexity} \label{proof:kernel_complexity}

From definition \ref{def:1}, given any perturbation $\widetilde{\mathcal{P}}$ on $\mathcal{P}$, there must exist a compact subset $K_i \subset U_i$ s.t.  $\left.R\right|_{\mathcal{P} \setminus \textrm{int}(K_i)} = \left.\textrm{id}\right|_{\mathcal{P} \setminus \textrm{int} (K_i)}$ and $\left.R\right|_{\textrm{int}(K_i)} \neq \left.\textrm{id}\right|_{\textrm{int}(K_i)}$. Furthermore there exists a linear homeomorphism between an open subset $\widetilde{U_i} \subseteq U_i$ with $\mathbb{R}^{d-1}$ which contains $K_i$.

We parametrize $K_i$ using a smooth function $g: \mathcal{T} \rightarrow  K_i \:$ s.t. $g(t) \in \partial K_i \; \forall t \in \partial \mathcal{T}$.

We further define $g_\epsilon(t) = g(t) + \epsilon \eta (t)$, for some perturbation $\epsilon \in \mathbb{R}$ and a smooth function $\eta: \mathcal{T} \rightarrow  \mathbb{R}^{d-1}$ 
We also restrict $\eta$ such that $\eta(t) = \mathbf{0} \; \forall t \in \partial \mathcal{T}$ and $\exists \; t_0 \in \mathcal{T}$ s.t. $\eta(t_0) \neq g(t_0)$.
In other words, $\eta$ is a smooth function where $g_\epsilon(t) = g(t) \; \forall \epsilon>0, \forall t \in \partial \mathcal{T}$, but is not identical to $g$ for all $t \in \mathcal{T}$. Using $g_\epsilon(t)$, we define the manifold $\mathcal{P}_\epsilon = \{g_\epsilon(t):t \in \mathcal{T}\}$.

To complete the proof, we want to show that the kernel similarity between any two given points $x,y \in \mathbb{R}^D$ is lower when using the manifold $\mathcal{P}_\epsilon$ for $\epsilon>0$ as opposed to the manifold $\mathcal{P}_0$. 
We therefore want to compare the two respective kernels $k_{\epsilon}(x,y)$ and $k_0(x,y)$.
Note that in this proof we consider the local effects of $\mathcal{P}$ on the kernel similarity through $\mathcal{P}_0$ and $\mathcal{P}_\epsilon$ exclusively, ignoring the manifold $\mathcal{P} \setminus U_0$
Using Euler-Lagrange, we can calculate a lower bound for $d_{\textrm{geo}}(g_\epsilon(t)), g_\epsilon(t')))$. In particular, for any $t,t' \in \mathcal{T}$, $d_{\textrm{geo}}(g_\epsilon(t)), g_\epsilon(t'))) \geq d_{\textrm{geo}}(g_0(t), g_0(t'))$.

\begin{equation}
    d_{\textrm{geo}}(g_\epsilon(t), g_\epsilon(t')) \geq d_{\textrm{geo}}(g_0(t), g_0(t'))
\end{equation}

\begin{equation}  \label{eq:inequality}
    \exp[-\lambda d_{\textrm{geo}}(g_\epsilon(t), g_\epsilon(t'))] \leq  \exp[-\lambda d_{\textrm{geo}}(g_0(t), g_0(t'))]
\end{equation}

\begin{equation} \label{eq:inequality2}
    \int_\mathcal{T} \int_\mathcal{T} \exp[-\lambda d_{\textrm{geo}}(g_\epsilon(t), g_\epsilon(t'))] \;  \textrm{d}t \textrm{d}t' \leq \int_\mathcal{T} \int_\mathcal{T} \exp[-\lambda d_{\textrm{geo}}(g_0(t), g_0(t'))] \;  \textrm{d}t \textrm{d}t'
\end{equation}

Note that in Eq. \eqref{eq:inequality2} we are integrating over all possible values of $t,t'$, therefore the inequality is tight iff $g_\epsilon(t) = g_0(t)$ $\forall t \in \mathcal{T}$; i.e. $\epsilon=0$ (see proof in \ref{proof:integral}). The case of $\epsilon = 0$ is trivial; we instead assume $\epsilon > 0$, in which case we can establish the following strict inequality: 

\begin{equation} \label{eq:inequality3}
    \int_\mathcal{T} \int_\mathcal{T} \exp[-\lambda d_{\textrm{geo}}(g_\epsilon(t), g_\epsilon(t'))] \;  \textrm{d}t \textrm{d}t' < \int_\mathcal{T} \int_\mathcal{T} \exp[-\lambda d_{\textrm{geo}}(g_0(t), g_0(t'))] \;  \textrm{d}t \textrm{d}t'
\end{equation}

Define uniform random variables $T$, $T'$ over the domain of $g$, i.e. $T, T' \sim \mathcal{U}_{\mathcal{T}}$. Then we have:

\begin{equation} \label{eq:inequality_int_over_t}
    \mathbb{E}_{T,T' \sim \mathcal{U}_{[0,1]}}[ \exp[-\lambda d_{\textrm{geo}}(g_\epsilon(T), g_\epsilon(T'))]] < \mathbb{E}_{T,T' \sim \mathcal{U}_{[0,1]}}[ \exp[-\lambda d_{\textrm{geo}}(g_0(T), g_0(T'))]]
\end{equation}

\begin{equation} \label{eq:inequalityfinal}
    \mathbb{E}_{M, M' \sim p_\epsilon(M)}[ \exp[-\lambda d_{\textrm{geo}}(M, M')]] < \mathbb{E}_{M, M' \sim p_0(M)}[ \exp[-\lambda d_{\textrm{geo}}(M, M')]]
\end{equation}

We define the random variable $M = g_\epsilon(T)$ with distribution $p_\epsilon(M)$. The distribution $p_\epsilon(M)$ represents the uniform distribution $\mathcal{U}_{\mathcal{T}}$ mapped to the manifold $\mathcal{P}_\epsilon$ using $g_\epsilon (T)$. The step from Eq. \eqref{eq:inequality_int_over_t} to Eq. \eqref{eq:inequalityfinal} uses a property of distribution transformations (Eq. 2.2.5 in \citet{casella_berger}).

Next, compare either side of Eq. \eqref{eq:inequalityfinal} to our kernel formulation shown below in Eq. \eqref{eq:inequalitykernel}. The kernel $k_\epsilon(x,y | \rho, \lambda)$ takes an expected value over $q_\epsilon(M|x,\rho)$ and $q_\epsilon(M'|y,\rho)$, which are equivalent to $p_\epsilon(M)$ and $p_\epsilon(M')$ weighted with respect to $x$, $y$, and a hyperparameter $\rho \geq 0$. 

\begin{align*} \numberthis \label{eq:inequalitykernel}
    k_\epsilon(x,y | \rho, \lambda) = \mathbb{E}_{M \sim q_\epsilon(M|x,\rho), M' \sim q_\epsilon(M'|y,\rho)}[\exp[-\lambda d_{\textrm{geo}}(M, M')]] \\
    s.t. \quad q_\epsilon(M | x, \rho) \propto  \exp[-\rho || x-M||_2^2] p_\epsilon(M)  \; \; \\
    s.t. \quad q_\epsilon(M' | y, \rho) \propto  \exp[-\rho || y-M'||_2^2] p_\epsilon(M') \\
\end{align*}

Note that when $\rho$ is set to zero, $q(M|x,0) = p(M)$ and $q(M'|y,0) = p(M')$. Therefore Eq. \eqref{eq:inequalityfinal} is equivalent to the inequality $k_\epsilon(x,y|0, \lambda) < k_0(x,y | 0, \lambda)$.

We next want to prove that the inequality $k_\epsilon(x,y | \rho, \lambda) < k_0(x,y|\rho, \lambda)$ also holds for non-zero values of $\rho$. For convenience, define

\begin{equation}
    f(\rho) = k_0(x,y|\rho, \lambda) - k_\epsilon(x,y | \rho, \lambda)
\end{equation}

Under this definition, we want to prove there exists $\rho_0>0$ such that $f(\rho) > 0$ $\forall \rho < \rho_0$. From Eq. \eqref{eq:inequalityfinal}, we established that $f(0)>0$. Assume that

\begin{equation}
\lim_{\rho \rightarrow 0} f(\rho) = c
\end{equation}

It therefore follows that $c>0$.
In addition, note that $f(\rho)$ is continuous with respect to $\rho$ (see proof in section \ref{proof:kernel_continuity}). Therefore for any $\epsilon > 0$ there exists $\delta > 0$ s.t. $\rho < \delta$ implies $|f(\rho) - c| < \epsilon$.

We choose $\epsilon = c$ and the define the corresponding $\delta$ to be $\rho_0$. Therefore:

\begin{equation}
    \rho < \rho_0 \Rightarrow |f(\rho) - c| < c
\end{equation}

\begin{equation}
    \rho < \rho_0 \Rightarrow  0 < f(\rho) < 2c
\end{equation}

Since this result holds for any $i$, it follows that the piecewise linear manifold $\mathcal{P}$ is a local minimum under any perturbation along a specific chart or combination of charts with respect to the kernel similarity $k(x,y) \; \forall x,y \in \mathbb{R}^D$.

\subsubsection{Proof: Inequality Tightness} \label{proof:integral}
From Eq. \eqref{eq:inequality2} to  Eq. \eqref{eq:inequality3}, we want to prove:

\begin{align*} \numberthis \label{eq:integral_proof1}
    \int_\mathcal{T} \int_\mathcal{T} \exp[-\lambda d_{\textrm{geo}}(g_\epsilon(t), g_\epsilon(t'))] \;  \textrm{d}t \textrm{d}t' = \int_\mathcal{T} \int_\mathcal{T} \exp[-\lambda d_{\textrm{geo}}(g_0(t), g_0(t'))] \;  \textrm{d}t \textrm{d}t' \\
    \Rightarrow g_\epsilon(t) = g_0(t) \quad \forall t \in \mathcal{T}
\end{align*}

Consider the LHS of Eq. \eqref{eq:integral_proof1}:

\begin{equation}
    \int_\mathcal{T} \int_\mathcal{T} \exp[-\lambda d_{\textrm{geo}}(g_\epsilon(t), g_\epsilon(t'))] \;  \textrm{d}t \textrm{d}t' = \int_\mathcal{T} \int_\mathcal{T} \exp[-\lambda d_{\textrm{geo}}(g_0(t), g_0(t'))] \;  \textrm{d}t \textrm{d}t' 
\end{equation}

\begin{equation} \label{eq:integral_proof2}
    \int_\mathcal{T} \int_\mathcal{T} \underbrace{\exp[-\lambda d_{\textrm{geo}}(g_0(t), g_0(t'))] -\exp[-\lambda d_{\textrm{geo}}(g_\epsilon(t), g_\epsilon(t'))]}_{h(t,t')} \;  \textrm{d}t \textrm{d}t' = 0
\end{equation}

Define $h(t,t')$ as the function inside the integrals in Eq. \eqref{eq:integral_proof2}. From Eq. \eqref{eq:inequality}, $h(t,t') \geq 0$ $\forall t,t' \in \mathcal{T}$. Since $h$ is continuous (see proof in \ref{proof:h_continuity}) and $\int_\mathcal{T} \int_\mathcal{T} h(t,t') \textrm{d}t \textrm{d}t' = 0$, it follows that $h(t,t') = 0$ $\forall t,t' \in \mathcal{T}$ (Ch.6 \citet{baby_rudin}).

It therefore follows that:

\begin{equation}\label{eq:integral_proof3}
    \exp[-\lambda d_{\textrm{geo}}(g_0(t), g_0(t'))] = \exp[-\lambda d_{\textrm{geo}}(g_\epsilon(t), g_\epsilon(t'))] \quad \forall t,t' \in \mathcal{T}
\end{equation}

From the definition of $\eta(t)$ in $g_\epsilon(t) = g(t) + \epsilon \eta(t)$, there must exist $t \in \mathcal{T}$ s.t. $\eta(t) \neq 0$. Therefore $\epsilon$ must be zero for Eq. \eqref{eq:integral_proof3} to hold. It follows that $g_\epsilon(t) = g_0(t)$ $\forall t \in \mathcal{T}$.

\subsubsection{Proof: Continuity of \texorpdfstring{$h(t,t')$}{h(t,t')}} \label{proof:h_continuity}

We prove that $h(t,t')$ is continuous with respect to $t,t'$. First note that by definition, $g_\epsilon(t)$ is a continuous parametrization of the manifold $\mathcal{P}_\epsilon$. From \citet{burago2022course}, it follows that for any two points $g_\epsilon(t), g_\epsilon(t') \in \mathcal{P}_\epsilon$,  $d_{\textrm{geo}}(g_\epsilon(t),g_\epsilon(t'))$ is continuous. Since the exponential functional preserves continuity and the sum of continuous functions are also continuous, it follows that $h(t,t')$ is continuous.

\subsubsection{Proof: Continuity of \texorpdfstring{$k(x,y)$}{k(x,y)} With Respect To \texorpdfstring{$\rho$}{Rho}} \label{proof:kernel_continuity}

We prove that $k(x,y)$ is continuous with respect to $\rho$.
\begin{equation}
k(x,y) = \int \int \exp[-\lambda d_{\textrm{geo}}(m, m')] \; q(m|x, \rho) \; q(m'|y, \rho) \; \textrm{d} m \textrm{d} m'
\end{equation}

\begin{align*} \numberthis \label{eq:continuousproof}
    =\frac{1}{Z_m(\rho) Z_{m'}(\rho)}  \int \int \mathcal{A} \underbrace{\exp[-\rho (|| x-m||_2^2 + || y-m'||_2^2)]}_{Z(\rho)} \; \textrm{d} m \textrm{d} m' \\ 
    s.t. \quad Z_m(\rho) = \int \exp[-\rho || x-m||_2^2] p(m) \; \textrm{d}m \; \; \; \;  \\ 
    \quad Z_{m'}(\rho) = \int \exp[-\rho || y-m'||_2^2] p(m') \; \textrm{d}m' \\
    \mathcal{A} = \exp[-\lambda d_{\textrm{geo}}(m, m')] p(m)p(m') \; \; \; \; \; \; \; \quad 
\end{align*}

Define $h(\rho) = \rho \mathcal{B}$, where $\mathcal{B}$ is a constant. Consider $h(\rho) - h(\rho_0)$, where $\rho_0$ is a fixed positive constant:

\begin{equation}
    |h(\rho) - h(\rho_0)| = |\rho \mathcal{B} - \rho_0 \mathcal{B}|
\end{equation}

\begin{equation}
    = |(\rho - \rho_0)\mathcal{B}| < \delta |\mathcal{B}|
\end{equation}

It follows that $\forall \; \epsilon>0$, $\exists \; \delta = \frac{\epsilon}{|\mathcal{B}|} > 0$ such that $|\rho - \rho_0| < \delta \; \Rightarrow \; |h(\rho) - h(\rho_0)| < \epsilon$. Therefore $h$ is continuous for all $\rho \in \mathbb{R}^+$. 

We set $\mathcal{B}$ to be $|| x-m||_2^2$, $|| y-m'||_2^2$, and $|| x-m||_2^2 + || y-m'||_2^2$, which shows that $Z_m (\rho)$, $Z_{m'}(\rho)$, and $Z(\rho)$ are also continuous, respectively. It then follows that the entirety of Eq. \eqref{eq:continuousproof} is continuous. 

%%%%%%%%%%%%%%%%%%%%%%%%%%%%%%%%%%%%%%%%%%%%%%%%%%%%%
%%%%%%%%%%%%%%%%%%%%%%%%%%%%%%%%%%%%%%%%%%%%%%%%%%%%%
%%%%%%%%%%%%%%%%%%%%%%%%%%%%%%%%%%%%%%%%%%%%%%%%%%%%%
%%%%%%%%%%%%%%%%%%%%%%%%%%%%%%%%%%%%%%%%%%%%%%%%%%%%%

\subsection{Extending to Multiclass Classifiers} \label{sec:multiclass}
In the multiclass case we define a black-box prediction model $F:\mathcal{X} \to \mathbb{R}^c$. We consider the one-vs-all DB for every class $y \in \mathcal{Y} = \{1, \ldots, c\}$, defined as $\mathcal{M}_{F_y} = \{x \in \mathbb{R}^{D} : F_y(x) = \max_{i \in \mathcal{Y}} \: F_i(x) = \max_{j \neq y \in \mathcal{Y}} F_j(x) \}$, where $F_y$ indicates the model output for class $y$. We then apply the GPEC framework separately to each class using the respective DB. The uncertainty estimate of the GP model would be of dimension $d \times c$.

\subsection{Feature Dependency in GPEC Output} \label{sec:vector_valued_GP}

A vector-valued GP is an extension of the traditional GP which has vector-valued output.
Let $\mathcal{X} \subseteq \mathbb{R}^D$ be the data space and $E: \mathcal{X} \rightarrow \mathbb{R}^D$ be an explainer with explanations $e = E(x) \; \forall x \in \mathcal{X}$. We sample $\mathcal{X}$ and generate $N$ pairs $\mathcal{S} = \{(x_1,e_1), ..., (x_N,e_N)\}$
In the main text, we train a vector-valued GP on each explanation dimension independently; i.e. we train $D$ independent scalar-valued GPs. This approach has the advantage of simplicity and implementation efficiency. However, alternative approaches can be used to enforce \emph{a priori} dependency between the dimensions of the vector-valued GP output.
Many matrix-valued kernels for vector-valued GPs have been investigated (see \citet{kernels_for_vvf} for a review). In particular, we review \emph{separable} kernels below, which allow an intuitive decomposition for the matrix-valued kernel.

Let $\mathbb{N}_{D} = \{z \in \mathbb{N}: z \leq D\}$. Let $k:\mathcal{X} \times \mathcal{X} \rightarrow \mathbb{R}_{\geq 0}$ and $k_T:\mathbb{N}_{D}\times \mathbb{N}_{D} \rightarrow \mathbb{R}_{\geq 0}$ be a scalar-valued kernel functions.
The function $k$ represents the standard kernel function for a GP (e.g. the WEG kernel for GPEC). The function $k_T$ represents an encoded similarity between tasks\footnote{To avoid confusion, we adopt the terminology of multi-task learning and refer to each explanation dimension $\{1,...,D\}$ as \emph{tasks}.} $\{1,...,D\}$. In the context of multi-task learning, $k$ and $k_T$ are sometimes referred to as the \emph{base kernel} and \emph{task kernel}, respectively.
We next define the respective kernel matrices $K$ and $B$.

\begin{equation}
    K = \left[ \begin{array}{ccc}
        k(x_1,x_1) & \dots & k(x_1,x_N) \\
        \vdots & \ddots &  \vdots   \\
        k(x_N,x_1)  & \dots & k(x_N,x_N)   \\
        \end{array} \right] \quad \quad
    B = \left[ \begin{array}{ccc}
        k_T(1,1) & \dots & k_T(1,D) \\
        \vdots & \ddots &  \vdots   \\
        k_T(D,1)  & \dots & k_T(D,D)   \\
        \end{array} \right]
\end{equation}

We can then define the block matrix $R = B \otimes K$, where $\otimes$ represents the Kronecker product.
We define the class of kernels which can be written in such a form as \emph{separable kernels}.

\begin{equation}
\small
    R = \left[ \begin{array}{ccc}
        \left[ \begin{array}{ccc}
        k(x_1,x_1)k_T(1,1) & \dots & k(x_1,x_N)k_T(1,1) \\
        \vdots & \ddots &  \vdots   \\
        k(x_N,x_1)k_T(1,1)  & \dots & k(x_N,x_N)k_T(1,1)   \\
        \end{array} \right]
        
        & \dots & \dots \\
        \vdots & \ddots &  \vdots   \\
        \dots  & \dots &
        
        \left[ \begin{array}{ccc}
        k(x_1,x_1)k_T(D,D) & \dots & k(x_1,x_N)k_T(D,D) \\
        \vdots & \ddots &  \vdots   \\
        k(x_N,x_1)k_T(D,D)  & \dots & k(x_N,x_N)k_T(D,D)   \\
        \end{array} \right]
        
        \\
        \end{array} \right]
\end{equation}

The kernel matrix $R$ of dimension $ND \times ND$ can therefore be used in the vector-valued GP.
In the simple case, where we assume independent output, we can set $B$ to be the identity matrix (i.e. $k_T(i,j) = \delta_{ij} \; \forall i,j \in \mathbb{N}_D$, where $\delta$ is the Kronecker delta). This is the case we assume in the GPEC formulation in the main text.
However, we can alternatively encode relatedness between outputs by selecting an appropriate $k_T$.

For example, \citet{sheldon2008graphical} define a user-defined adjacency matrix of a graph where the nodes of the graph represent tasks, and the edges represent task similarity. This allows the user to encode \emph{a priori} relationships between outputs. Alternatively, \citet{learning_multiple_tasks} define $k_T$ based on clustering, and enforce within-cluster similarity for tasks.

Any such methods can be used with GPEC by setting the base kernel $k$ to be the WEG kernel defined in Section \ref{sec:weighted_geodesic_kernel} and then selecting the desired task kernel $k_T$.

\section{IMPLEMENTATION DETAILS} \label{app:implementation_details}

\subsection{Adversarial Sample Filtering for Multi-class Models}
\label{app:adv}
Following \citet{karimi2019characterizing}, we elect to sample from multi-class neural network decision boundaries by using a binary search algorithm on pairs of adversarial samples.
Specifically, given a test-point $x_0 \in \mathbb{R}^D$ and model prediction $y = \text{argmax}_{k \in \mathcal{Y}} F(x_0)$, decision boundary points may be generated by the following procedure: 

First, for each class $v \in \mathcal{Y}$ a set of $M_v$ points is randomly sampled from the set of train points on which the model predicts class $v$: 
\begin{equation}
 \mathcal{X}_v \subseteq
\{ x : \text{argmax}_{k \in \mathcal{Y}} \ F(x) = v, \ x \in \mathcal{X}_{tr} \}, \ |\mathcal{X}_v| = M_v    
\end{equation}
$\forall v \in \mathcal{Y}$.
An untargeted adversarial attack using a given $l_p$ norm and radius $\epsilon$ is generated for each point in $\mathcal{X}_y$, the set of points with the same class prediction as $x_0$.
Each attack output $Attack_{U}(x, \epsilon) \in \mathbb{R}^D$ is paired with its corresponding input, resulting in the set 
\begin{equation}
\mathcal{X}_{y'} = \{(x, Attack_{U}(x, \epsilon)): x \in \mathcal{X}_y \},
\end{equation}
where for an element $(a,b) \in \mathcal{X}_{y'}$ we have $ \text{argmax}_{k \in \mathcal{Y}} \ F(a) = y, \text{argmax}_{k \in \mathcal{Y}} \ F(b) = v \neq y$, where $v$ is an unspecified class. 

Likewise, a targeted adversarial attack, with target class $y$, is run on each point in each of the sets of points that are not predicted as class $y$.
Each attack output $Attack_y(x, \epsilon) \in \mathbb{R}^D$ may be paired with its input $x$ resulting in sets 
\begin{equation}
\mathcal{X}_{v'} = \{ (x, Attack_y(x, \epsilon)) : x \in \mathcal{X}_v \}
\end{equation} 
$\forall v \neq y \in \mathcal{Y}$. Here, for an element $(a,b) \in \mathcal{X}_{v'}$ we have $\text{argmax}_{k \in \mathcal{Y}} \ F(a) = v, \text{argmax}_{k \in \mathcal{Y}} \ F(b) = y$. 

%%% maybe change to bigcup style later replace we have generated with bigcup 
Thus, we have generated a diverse set of $\sum_{v \in \mathcal{Y}}M_v$ pairs of points that lie on opposite sides of the decision boundary for class $y$.
The segment between any pair from a given set $\mathcal{X}_{v'} \ v \neq y$ will necessarily contain a point on the class $v$ v.s. class $y$ decision boundary.
Likewise, in the interest of further diversity, segments between any pair from the set $\mathcal{X}_{y'}$ will contain a point on the class $v$ v.s. class $y$ decision boundary, where $v \neq y \in \mathcal{Y}$ is unspecified.
A binary search may be applied to each pair of samples to find the boundary point. 

In practice, the entire procedure may be applied for all classes as a single post-processing step immediately after training.
The results may be saved as a dictionary of boundary points which may be efficiently queried via the model predicted class of any given test point. 

In our implementation, each adversarial attack is attempted multiple times, once using each radius value $\epsilon$ in the list: $[0.0, 2e^{-4}, 5e^{-4}, 8e^{-4}, 1e^{-3}, 1e^{-3},1.5e^{-3},2e^{-3},3e^{-3},1e^{-2},1e^{-1},3e^{-1},5e^{-1},1.0]$.
For a given input, the output of the successful attack with smallest $\epsilon$ is used. If no attack is successful at any radius, the input is discarded from further consideration. 
We apply Projected Gradient Descent (PGD) \citep{madry2018towards} attacks with the $l_\infty$ norm for both targeted and untargeted attacks, using the implementation provided in \citet{rauber2017foolbox, rauber2017foolboxnative}.  The  $M_c$ values used for the relevant datasets are indicated below in  Appendix \ref{app:dataset_details}. 

\subsection{Geodesic Distance Approximation}
We utilize the ISOMAP algorithm \cite{tenenbaumGlobalGeometricFramework2000} to approximate geodesic distances. We adapt the code\footnote{\url{https://github.com/redst4r/riemannian_latent_space}} from \cite{fast_geodesics} in our implementation. Given a set of samples from a manifold, ISOMAP constructs a graph where each sample is node. Graph edges are populated by Euclidean distances between samples. After defining the graph, we use a shortest path length algorithm from NetworkX \cite{networkx} to approximate geodesic distance.

\section{EXPERIMENT SETUP}

\subsection{Datasets and Models} \label{app:dataset_details}

%# TODO: Add model information
\textbf{Census.} The UCI Census dataset consists of 32,561 samples from the 1994 census dataset. Each sample is a single person's response to the census questionaire. An XGBoost model is trained using the 12 features to predict whether the individual has income $\geq$ \$50k.
    
 \textbf{Online Shopper.} The UCI Online Shoppers dataset consists of clickstream data from 12,330 web sessions. Each session is generated from a different individual and specifies whether a revenue-generating transaction takes place. There are 17 other features including device information, types of pages accessed during the session, and date information. An XGBoost model is trained to predict whether a purchase occurs.
    
\textbf{German Credit.} The German Credit dataset consists of 1,000 samples; each sample represents an individual who takes credit from a bank. The classification task is to predict whether an individual is considered a good or bad risk. Features include demographic information, credit history, and information about existing loans. Categorical features are converted using a one-hot encoding, resulting in 24 total features.
    
\textbf{MNIST.} The MNIST dataset \citep{lecun-mnisthandwrittendigit-2010} consists of 70k grayscale images of dimension 28x28. Each image has a single handwritten numeral, from 0-9. A fully connected network with layer sizes 784-700-400-200-100-10 and ReLU activation functions was trained and validated on on 50,000 and 10,000 image label pairs, respectively. Training lasted for $30$ epochs with initial learning rate of $2$ and a learning rate decay of $\gamma = 0.5$ when training loss is plateaued. During adversarial example generation we used $M_y = 500$ and $M_c = 50 \ \forall c \neq y$.

\textbf{Fashion MNIST.} The Fashion MNIST dataset \citep{fmnist} contains 70,000 grayscale images of dimension 28x28. There are 10 classes, each indicating a different article of clothing. We train a MLP model with the same architecture used for the MNIST dataset, however we increase training to $100$ epochs and increase the initial learning rate to $3$. During adversarial example generation we used $M_y = 500$ and $M_c = 50 \ \forall c \neq y$.

\textbf{CIFAR10} The CIFAR10 dataset \citep{cifar10} contains 60,000 color images of dimension 32x32. Each image contains an object from one of 10 classes: airplane, automobile, bird, cat, deer, dog, frog, horse, ship, truck. We train a Resnet-18 model \citep{resnet} for use in the experiments. During adversarial example generation we used $M_y = 500$ and $M_c = 50 \ \forall c \neq y$.

\subsection{Competitor Implementation Details} \label{app:competitor_details}

\textbf{BayesLIME and BayesSHAP.} \citet{reliable:neurips21} extend the methods LIME and KernelSHAP to use a Bayesian Framework. BayesLIME and BayesSHAP are fit using Bayesian linear regression models on perturbed outputs of the black-box model. The posterior distribution of the model weights are taken as the feature attributions instead of the frequentist estimate that characterizes LIME and KernelSHAP. We take the expected value of the posterior distribution as the point estimate for feature attributions, and the 95\% credible interval as the estimate of uncertainty. To implement BayesLIME and BayesSHAP we use the public implementation\footnote{\url{https://github.com/dylan-slack/Modeling-Uncertainty-Local-Explainability}}. We set the number of samples to 200, disable discretization for continuous variables, and calculate the explanations over all features. Otherwise, we use the default parameters for the implementation.

\textbf{CXPlain.} \citet{cxplain} introduces the explanation method CXPlain, which trains a surrogate explanation model based on a causal loss function. After training the surrogate model, the authors propose using a bootstrap resampling technique to estimate the variance of the predictions. In our experiments we implement the publicly available code\footnote{\url{https://github.com/d909b/cxplain}}. We use the default parameters, which include using a 2-layer UNet model \cite{unet} for the image datasets and a 2-layer MLP model for the tabular datasets. We take a 95\% confidence interval from the bootstrapped results as the estimate of uncertainty.

\subsection{Regularization Parameter Overview} \label{app:regularization_parameter}
\textbf{$\boldsymbol{L_2}$ regularization}. Let $f$ be a neural network with parameters $\hat \theta = \arg \min_{\theta} \ell(y, f_\theta) + \lambda||\theta||^2_2$, where $\ell$ is a given loss function. The component $\lambda||\theta||^2_2$ adds a penalty for the magnitude of the parameters $\theta$, which is controlled by parameter $\lambda$. In our experiments we increase $\lambda$ to increase the regularization of the model $f$. Increasing $\lambda$ encourages the model to have smaller values of $\theta$, which results in a lower complexity model. We can see that $\lim_{\lambda \rightarrow \infty} \arg \min_{\theta} \ell(y, f_\theta) + \lambda||\theta||^2_2$ becomes the zero vector, which implies that the model becomes linear.

\textbf{Softplus $\boldsymbol{\beta}$.} For a given neural network $f$ with ReLU activation functions, we replace the ReLU functions with the Softplus function: $\textrm{Softplus}(x;\beta)=\frac{1}{\beta} \log(1+\exp(\beta x))$. The Softplus function is a smooth approximation of ReLU, which has been previously investigated \citep{dombrowski2019explanations, wang2020smoothed} in the context of improving neural network smoothness. Smoothness regularizers have been shown to reduce the complexity of neural networks and improve generalization \citep{rosca2020case}. Decreasing $\beta$ increases the smoothing effect of the Softplus, which reduces the complexity of the model.

\textbf{$\boldsymbol{\gamma}$ parameter for XGBoost.} We increase the $\gamma$ parameter in the XGBoost loss function (Eq. (2) in \citet{xgboost}):
$$
\begin{aligned}
& \mathcal{L}(\phi)=\sum_i l\left(\hat{y}_i, y_i\right)+\sum_k \Omega\left(f_k\right) \\
& \text { where } \Omega(f)=\gamma T+\frac{1}{2} \lambda\|w\|^2
\end{aligned}
$$

The $\Omega(f_k)$ component regularizes the complexity of the XGBoost model, which is an ensemble of functions $f_k$. The $\gamma$ parameter penalizes the magnitude of $T$, which represents the number of leaves in each tree. Reducing the number of leaves in each tree function corresponds to smaller trees with less complexity. Therefore, increasing $\gamma$ results in XGBoost models reduces the overall complexity of the model.

\section{ADDITIONAL RESULTS} \label{app:additional_results}

In Section \ref{app:execution_time} we show an execution time comparison for explanation uncertainty methods. In Section \ref{app:diabetes} we perform a case study using GPEC on a diabetes prediction task. In Section \ref{app:illustrative_examples} we show illustrative examples from MNIST, f-MNIST, and CIFAR10. In Section \ref{app:sensitivity_analysis} we evaluate the sensitivity of GPEC to changes in parameters $\rho$ and $\lambda$. In Section \ref{app:addnl_results_fullfat} we perform an experiment evaluating function approximation uncertainty component of GPEC.
In Section \ref{app:addnl_results_regularization} we report standard error and parameters for the regularization experiment in Section \ref{sec:experiment_regularizationtest}.
In Section \ref{app:addnl_results_uncertainty_visualization} we show additional results from the Uncertainty Visualization Experiment.

\subsection{Execution Time Results} \label{app:execution_time}
\begin{table*}[t]
    %------------------------------------------
    % \setlength{\belowcaptionskip}{-15pt}

    \scriptsize
    \setlength{\tabcolsep}{3.0pt}
    \renewcommand{\arraystretch}{1.2}
    \centering
    \begin{tabular}{| l | c | c | c | c | c | c |}
        %--------------
        \hline
        & 
        \multicolumn{1}{c|}{Census}  & 
        \multicolumn{1}{c|}{Online Shoppers} &
        \multicolumn{1}{c|}{German Credit} &
        \multicolumn{1}{c|}{MNIST} &
        \multicolumn{1}{c|}{f-MNIST} &
        \multicolumn{1}{c|}{CIFAR10} \\
        \hline 

        %--------------
            
        GPEC-WEG &
            0.11 &
            0.37 &
            0.07 &
            12.90 &
            18.15 &
            1.77

            \\

        %--------------
        GPEC-RBF &
            0.00 &
            0.00 &
            0.02 &
            8.95 &
            7.41 &
            1.29

            \\

        %--------------
        CXPlain &
            0.05 &
            0.06 &
            0.04 &
            9.76 &
            18.18 &
            8.27

            \\
        %--------------
        BayesSHAP &
            140.40 &
            54.56 &
            4.86 &
            42,467 &
            42,361 &
            --

            \\

        %--------------
        BayesLIME &
            91.29 &
            54.60 &
            4.83 &
            41,832 &
            41,992 &
            --
            \\
    
        \hline
    \end{tabular}
    \caption{Inference time comparison (\emph{in seconds}) for estimating the uncertainty for all features for 100 samples. For MNIST and f-MNIST datasets, results represent execution time for calculating uncertainty estimates with respect to all ten classes. CIFAR10 results were calculated using a single A100 GPU; all other results were calculated using CPU only. CIFAR10 results for BayesSHAP and BayesLIME are omitted due to computational expense}
    \label{table:time_full}
    
    \end{table*}
    
\begin{table*}[t]
    %------------------------------------------
    % \setlength{\belowcaptionskip}{-15pt}

    \scriptsize
    \setlength{\tabcolsep}{3.0pt}
    \renewcommand{\arraystretch}{1.2}
    \centering
    \begin{tabular}{| l | c | c | c | c | c | c |}
        %--------------
        \hline
        & 
        \multicolumn{1}{c|}{Census}  & 
        \multicolumn{1}{c|}{Online Shoppers} &
        \multicolumn{1}{c|}{German Credit} &
        \multicolumn{1}{c|}{MNIST} &
        \multicolumn{1}{c|}{f-MNIST} &
        \multicolumn{1}{c|}{CIFAR10} \\
        \hline 

        %--------------
            
        GPEC \emph{(total)}&
            35.7 &
            22.5 &
            22.4 &
            94.4 &
            78.0 &
            254

            \\
            
        %--------------
        \hspace{2mm} Sample DB&
            35.7 &
            22.4 &
            22.4 &
            34.5 &
            27.8 &
            220

            \\
            
        %--------------
        % \hline
        Naive-GP &
            0.02 &
            0.02 &
            0.02 &
            7.03 &
            7.11 &
            1.29

            \\

        %--------------
        CXPlain &
            1.07 &
            0.34 &
            1.71 &
            3.54 &
            3.73 &
            3.85

            \\
            
        %--------------
        BayesSHAP &
            -- &
            -- &
            -- &
            -- &
            -- &
            --

            \\
        %--------------
        BayesLIME &
            -- &
            -- &
            -- &
            -- &
            -- &
            --

            \\

        \hline
    \end{tabular}
    \caption{Training time comparison (\emph{in minutes}) for various explanation methods. The ``Sample DB'' step for GPEC is included for clarity and indicates the execution time for drawing samples from the black-box DB (for all classes). CIFAR10 results were calculated using a single A100 GPU; all other datasets were calculated on CPU only. Note that BayesLIME and BayesSHAP methods do not have a training step.}
    \label{table:time_train}
    
\end{table*}

In Table \ref{table:time_full} we include inference time comparison for the methods implemented in this paper. Results are averaged over 100 test samples. For MNIST and f-MNIST datasets, results evaluate the time to calculate uncertainty estimates with respect to all classes. All experiments were run on an internal cluster using AMD EPYC 7302 16-Core processors. The CIFAR10 dataset was run using a single Nvidia A100 GPU; the other datasets were run on CPU only. 
We observe from the results that the methods that amortization methods (GPEC-WEG, GPEC-RBF, CXPlain) are significantly faster than perturbation methods BayesLIME and BayesSHAP.

In Table \ref{table:time_train} we include training time results for the implemented methods. The ``Sample DB'' step for GPEC is highlighted for clarity and indicates the execution time for drawing samples from the black-box DB. The GP regression model in GPEC can be retrained with different hyperparameter choices ($\rho$, $\lambda$) and/or training samples ($X$) using the same DB samples, therefore this step only needs to be performed once for each given dataset and black-box model combination.
This step is dependent on DB sampling algorithms (see App. \ref{app:adv}); improvements in these algorithms will decrease training time for GPEC.
Note that BayesLIME and BayesSHAP methods do not have a training step.

\subsection{Case Study: Diabetes Prediction with GPEC uncertainty} \label{app:diabetes}

\begin{figure}[t]
   \begin{center}
   \includegraphics[width=1\linewidth]{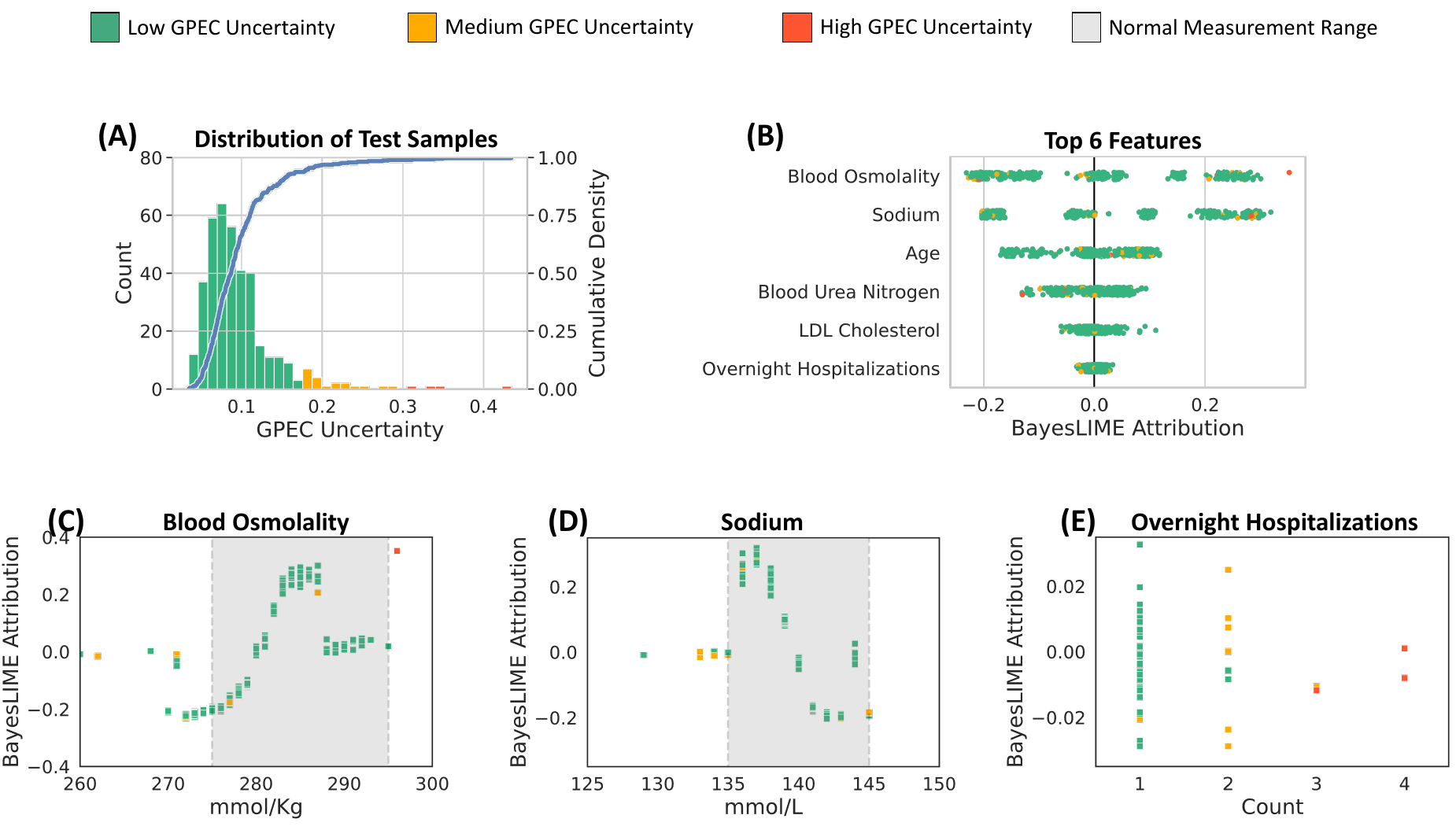}
   \end{center}
   \caption{Case study of using GPEC to improve understanding for diabetes prediction. \textbf{(A)} We categorize samples into tertiles for low, medium, and high GPEC uncertainty. \textbf{(B)} We plot the top features using BayesLIME attributions while also overlaying the GPEC uncertainty (green, yellow, red). \textbf{(C) - (E)} We visualize three top features by value, importance, and uncertainty. Patients with high GPEC uncertainty for a given feature may require further investigation due to function approximation uncertainty and DB-aware uncertainty.}
   \label{fig:nhanes}
\end{figure}

In this section we evaluate how GPEC can be used to improve understanding of model predictions and feature attributions.
We used the NHANES \citep{millerPlanOperationHealth1973} 2013-2014 dataset, which is an annual survey conducted by the Center for Disease Control and Prevention (CDC) and the National Center for Health Statistics (NHCS). It contains demographic, dietary, health exam, and survey data for 3,329 patients. We follow \citet{dinhDatadrivenApproachPredicting2019} in training an XGBoost model to predict the incidence of type-2 diabetes using a pre-selected set of 27 features. The model achieves an AUROC of $0.92$, which replicates the results in \citet{dinhDatadrivenApproachPredicting2019}. After training the model, we apply a feature attribution method, BayesLIME, to estimate the importance of each feature for a given patient's prediction of diabetes status. Since BayesLIME is a \emph{local} feature attribution method, different patients may have different features (e.g. lab tests, physical attributes) that may be indicative of diabetes.

To establish a confidence estimate that includes \emph{both} function approximation and DB-aware uncertainty, we apply GPEC over the test samples and categorize the samples in tertiles for low, medium, and high uncertainty (Fig. \ref{fig:nhanes}(A)).
In Figure \ref{fig:nhanes}(B) we plot the distribution of test samples for three features of high overall importance for the incidence of diabetes, as calculated using BayesLIME attributions. We observe that the majority of patients have low uncertainty.

In Figures \ref{fig:nhanes}(C) - (E), we investigate three top features to with respect to feature value, feature importance, and GPEC uncertainty.
In Figure \ref{fig:nhanes}(D), sodium, we see that there are 5 patients below the normal measurement range with medium uncertainty. For these patients, the BayesLIME attribution indicates that sodium level has minimal importance towards diabetes prediction, however the uncertainty score indicates that this explanation may not be reliable – more investigation is suggested.

In Figure \ref{fig:nhanes}(E), overnight hospitalizations, we see that having two hospitalizations is generally significant (higher magnitude of BayesLIME attribution), however these explanations have elevated uncertainty. The plot also indicates that having 3-4 hospitalizations generally has minimal impact on the prediction. This result is somewhat unexpected, and the high uncertainty suggests that these results should be investigated further. We hypothesize that there are relatively few patients with 2-4 overnight hospitalizations, leading to model overfitting and a higher GPEC uncertainty estimate.

\subsection{Illustrative Examples for MNIST, f-MNIST, and CIFAR10} \label{app:illustrative_examples}

\begin{figure}[t]
   \begin{center}
   \includegraphics[width=1\linewidth]{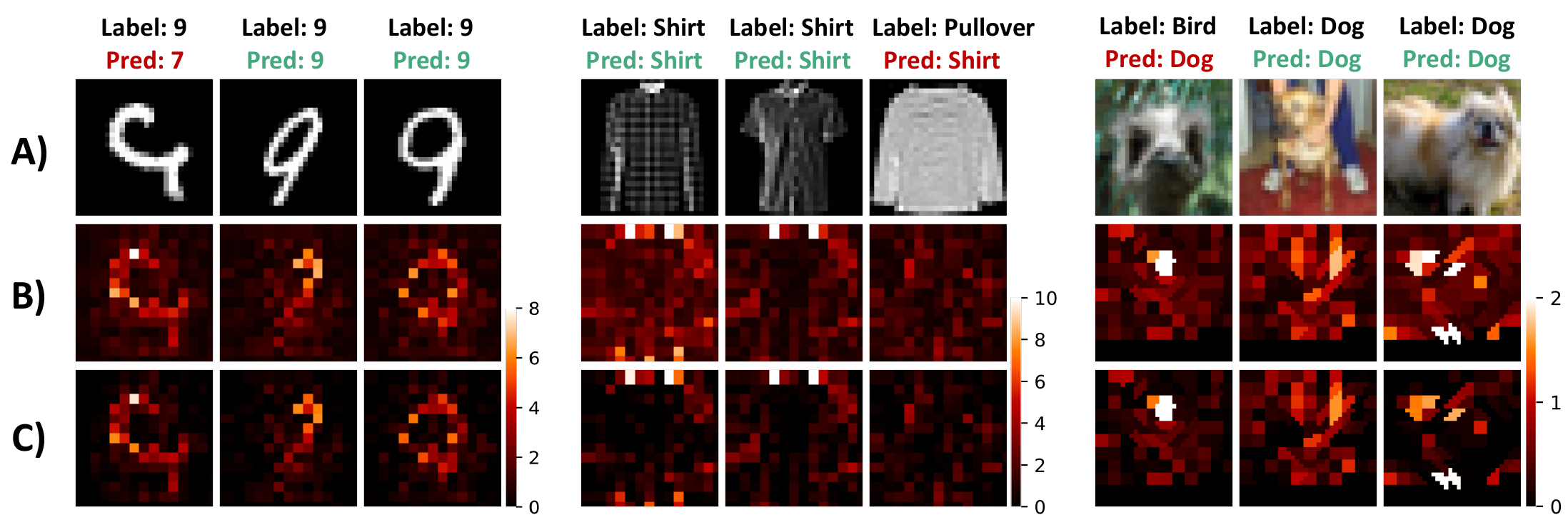}
   \end{center}
   \caption{Illustrative samples from MNIST (left), f-MNIST (middle), and CIFAR10 (right). Row (A) shows the original sample. Row (B) visualizes the upper bound of the confidence interval (CI) for feature attributions using GPEC uncertainty. Row (C) visualizes the corresponding lower bound of the CI. We use BayesLIME attributions with GPEC uncertainty to cacluate the intervals. SLIC superpixels are used to improve the interpretability of results.}
   \label{fig:illustrative_examples}
\end{figure}

We present illustrative examples from MNIST, f-MNIST, and CIFAR10 in Fig. \ref{fig:illustrative_examples}. For each image, we apply BayesLIME to generate feature attributions, then apply GPEC to estimate a confidence interval. To improve interpretability of the results, we use the \emph{simple linear iterative clustering} (SLIC) \citep{slic} method, which clusters similar pixels into \emph{superpixels}, and explain each superpixel rather than the individual pixels. The MNIST and f-MNIST datasetse us 196 superpixels, and the CIFAR10 dataset uses 96 superpixels. The upper and lower limit of the confidence interval is plotted in Fig. \ref{fig:illustrative_examples} row (B) and row (C). We observe that the GPEC confidence interval gives an estimate of uncertainty for the features for each image, which improves the interpretation of the feature attribution heatmap. The difference between upper and lower bounds is especially large for the background pixels in MNIST and f-MNIST datasets.

\subsection{Sensitivity Analysis of WEG Kernel Parameters} \label{app:sensitivity_analysis}

\begin{figure}[t]
   \begin{center}
   \includegraphics[width=0.6\linewidth]{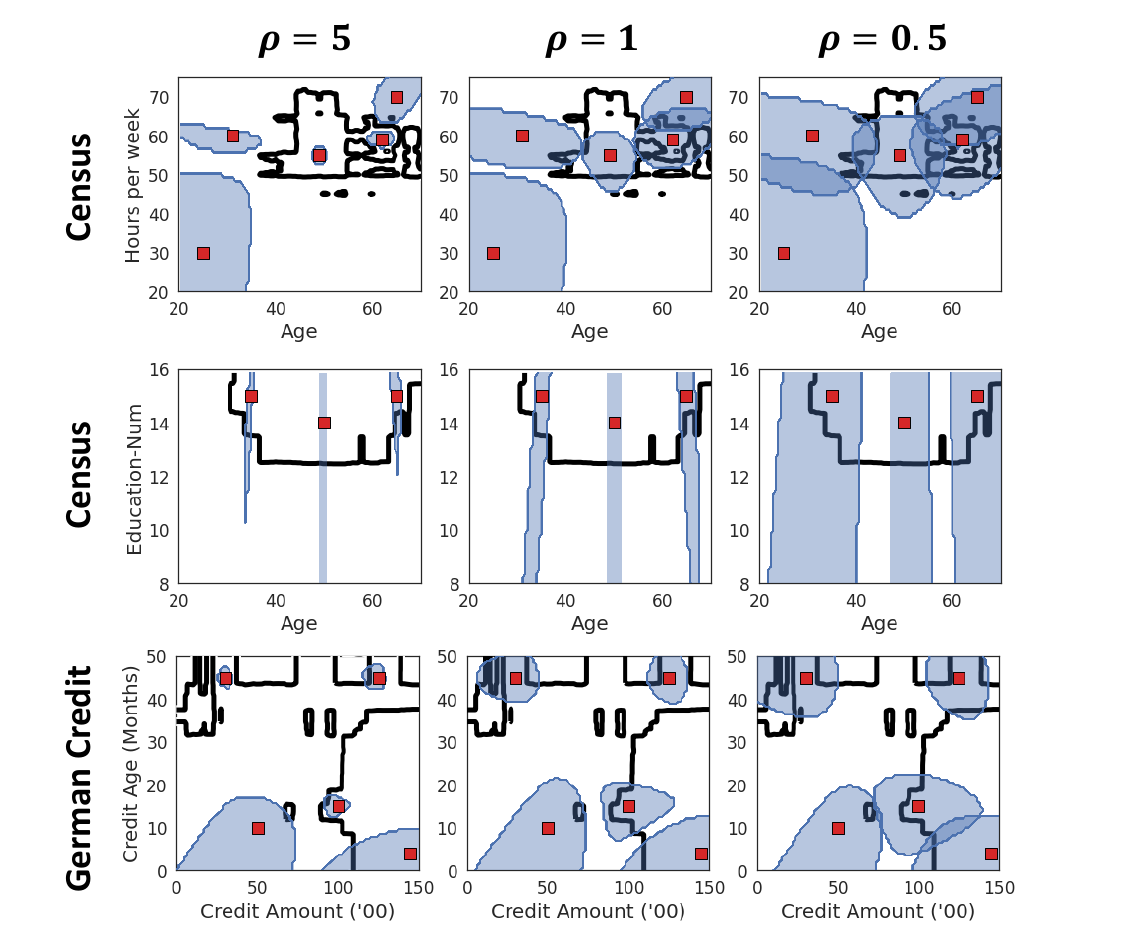}
   \end{center}
   \vspace{-4mm}
   \caption{Evaluation of the WEG kernel for various DBs and values of $\rho$. The black line indicates the DB for the associated black-box model. The blue region highlights the set $\{x' : k(x,x') \geq 0.9\}$ for a given $x$ (red). This region decreases in size when the local DB near $x$ becomes more complex. Increasing the hyperparameter $\rho$ increases the sensitivity of the WEG kernel similarity to DB complexity.}
   \label{fig:kernel_sensitivity_rho}
\end{figure}

\begin{figure}[t]
   \begin{center}
   \includegraphics[width=0.9\linewidth]{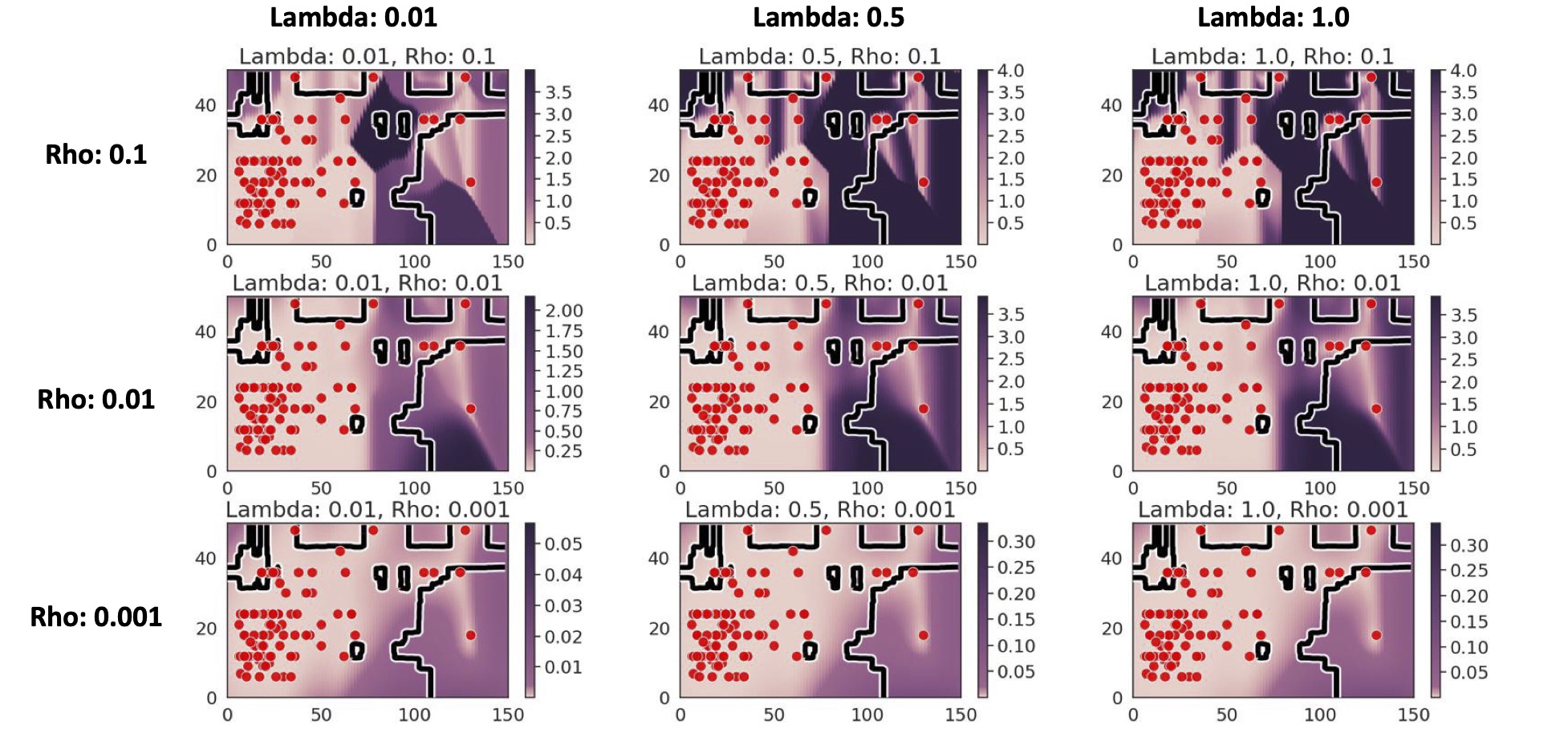}
   \end{center}
   \vspace{-4mm}
   \caption{Hyperparameter sensitivity analysis for the German Credit Dataset. Heatmap of estimated uncertainty for the x-axis variable under different $\rho$ and $\lambda$ parameter choices.}
   \label{fig:sensitivity_germancredit}
\end{figure}

\begin{figure}[t]
   \begin{center}
   \includegraphics[width=0.9\linewidth]{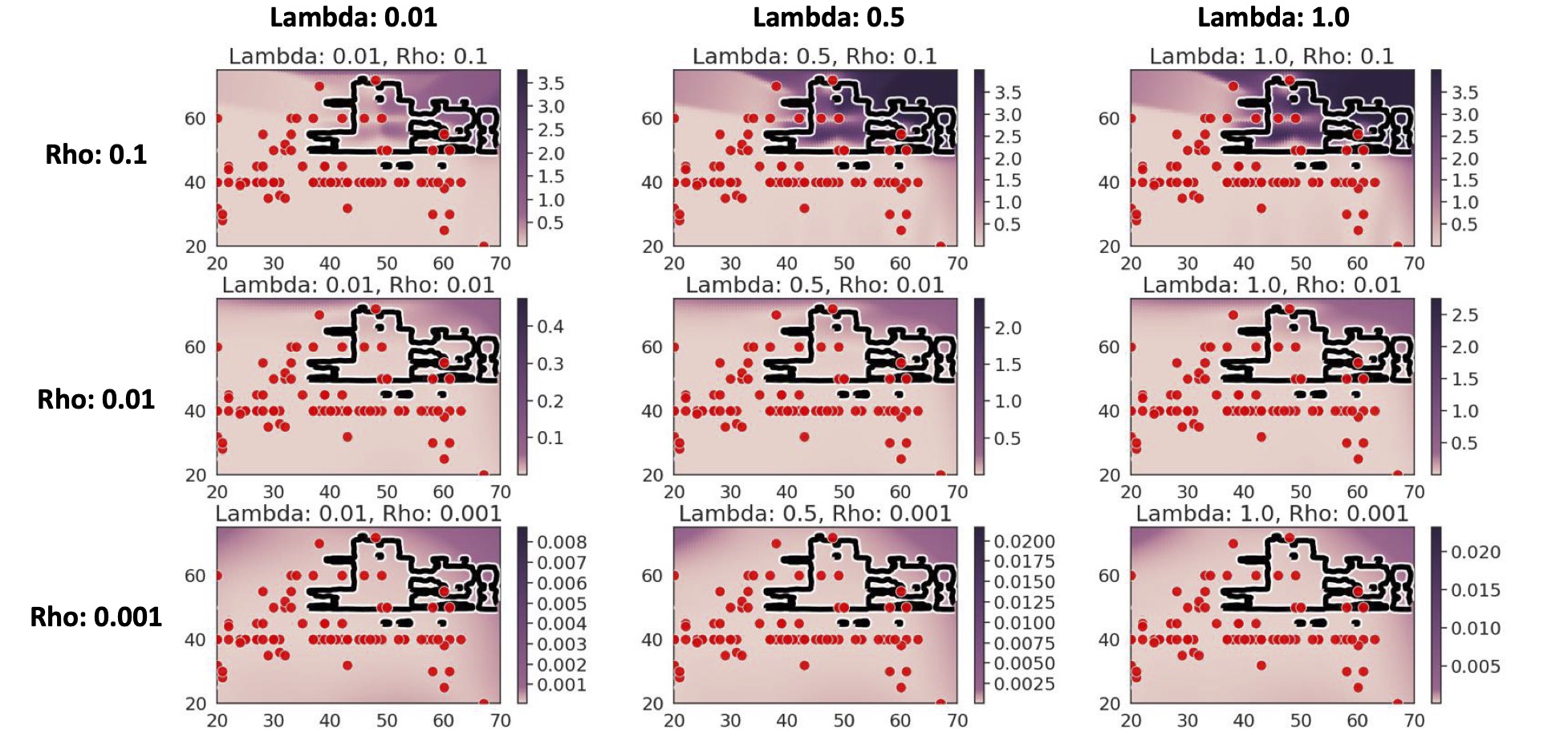}
   \end{center}
   \vspace{-4mm}
   \caption{Hyperparameter sensitivity analysis for the Census Dataset. Heatmap of estimated uncertainty for the x-axis variable under different $\rho$ and $\lambda$ parameter choices.}
   \label{fig:sensitivity_census1}
\end{figure}

\begin{figure}[t]
   \begin{center}
   \includegraphics[width=0.9\linewidth]{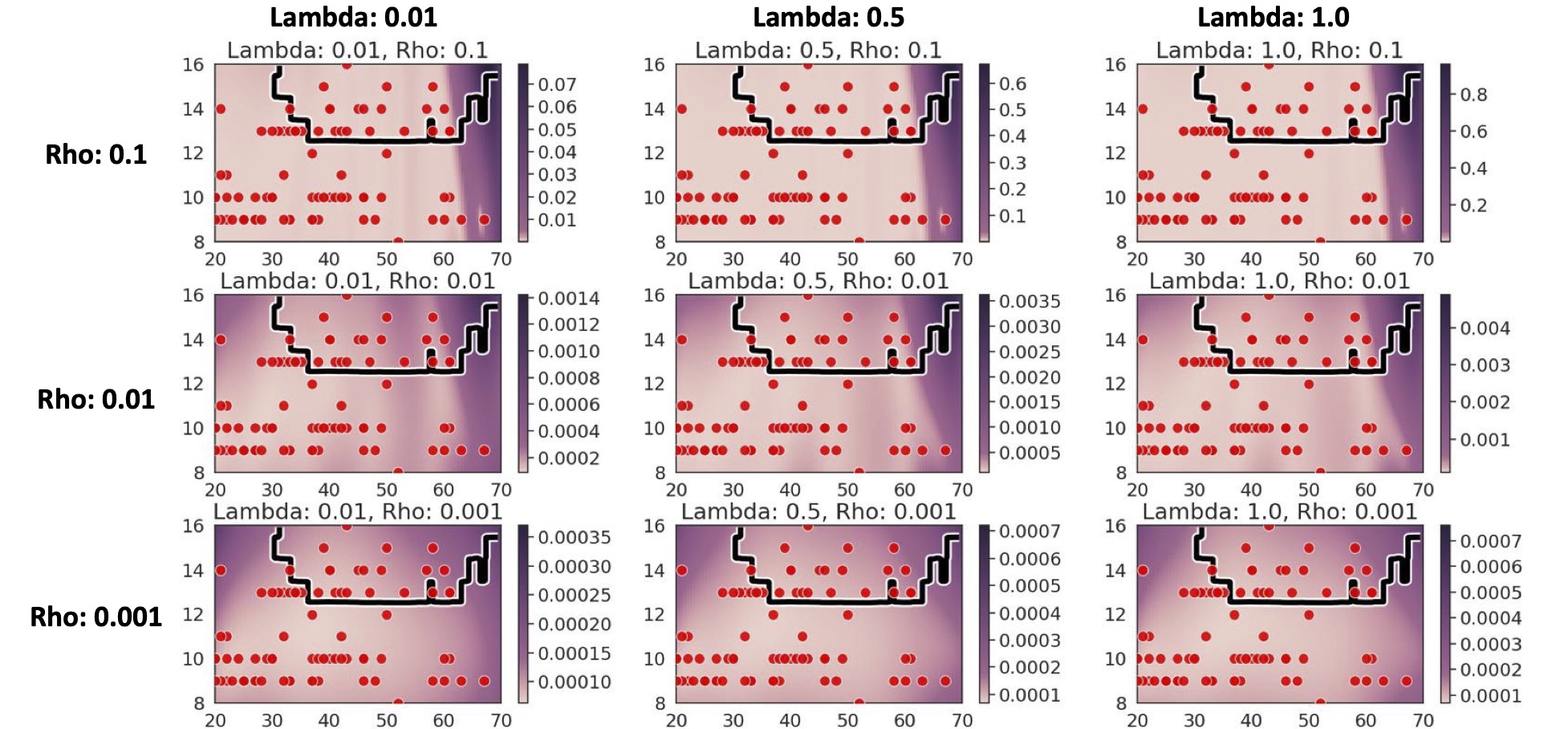}
   \end{center}
   \vspace{-4mm}
   \caption{Hyperparameter sensitivity analysis for the Census Dataset. Heatmap of estimated uncertainty for the x-axis variable under different $\rho$ and $\lambda$ parameter choices.}
   \label{fig:sensitivity_census2}
\end{figure}

The WEG kernel formulation uses two parameters, $\rho$ and $\lambda$. The parameter $\rho$ controls the weighting between each datapoint and the manifold samples. As $\rho$ increases, the WEG kernel places more weight on manifold samples close in $\ell_2$ distance to the given datapoint. The parameter $\lambda$ acts as a bandwidth parameter for the exponential geodesic kernel. Increasing $\lambda$ increases the effect of the geodesic distance along the manifold. Therefore decision boundaries with higher complexity will have an increased effect on the WEG kernel similarity.
Bayesian model selection methods such as log marginal likelihood maximization (see \cite{GPML}) can be used for selecting hyperparameters $\rho$ and $\lambda$.
In practice, it is also important to select $\lambda$ such that the EG kernel (Eq. \eqref{eqn:eg}) is positive-definite \citep{feragen_geodesic_exponential_kernel}, which can be identified through cross-validation.

 In Figure \ref{fig:kernel_sensitivity_rho} we extend the kernel similarity analysis in Figure \ref{fig:kernel_comparison}B to evaluate the WEG kernel for different DBs and $\rho$ values. We observe that the similarity-ball $\{x' : k(x,x') \geq 0.9\}$ (blue) for points near complex DB are generally smaller in size. Increasing $\rho$ increases sensitivity to nearby complex DB segments; i.e. values near complex DB segments will have correspondingly smaller similarity-balls for larger $\rho$.

In Figures \ref{fig:sensitivity_germancredit}, \ref{fig:sensitivity_census1}, and \ref{fig:sensitivity_census2} we plot heatmaps for various combinations of $\rho$ and $\lambda$ parameters to evaluate the change in the uncertainty estimate. The black line is the decision boundary and the red points are the samples used for training GPEC. Please note that the heatmap scales are not necessarily the same for each plot.

\subsection{Visualizing effects of Explainer Uncertainty in GPEC Estimate} \label{app:addnl_results_fullfat}

\begin{figure}[t]
   \begin{center}
   \includegraphics[width=1\linewidth]{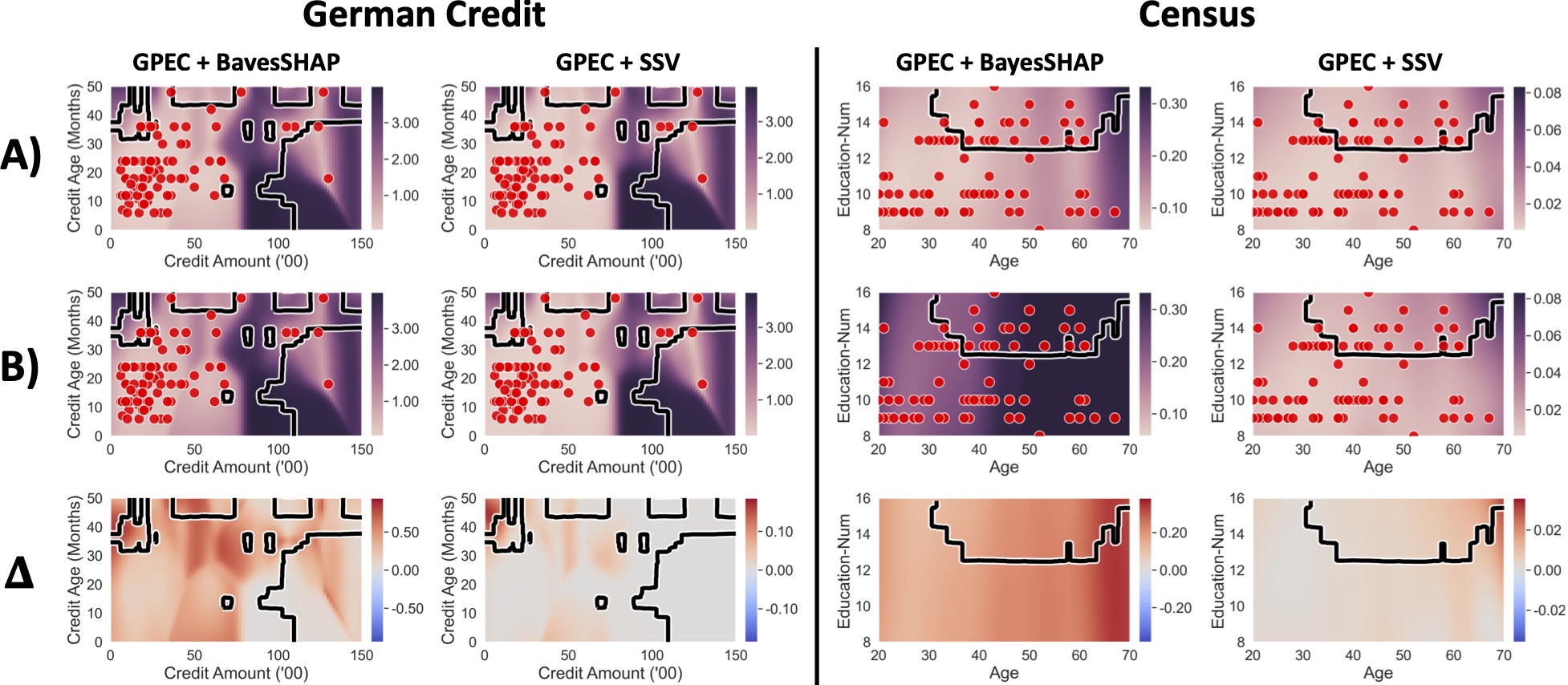}
   \end{center}
   \caption{Comparison of the change in quantified uncertainty of explanations as we change the number of samples for BayesSHAP and SSV. Row (A) visualizes the combined uncertainty estimate using GPEC and either BayesSHAP or SSV, using 200 samples for approximating the BayesSHAP / SSV explanation. In Row (B) we decrease the number of samples to 5 and recalculate the estimated uncertainty. Row ($\Delta$) represents the change in uncertainty estimate between (A) and (B). We see that the average uncertainty changes as we decrease the number of samples, which indicates that GPEC is able to capture the uncertainty arising from BayesSHAP / SSV approximation.}
   \label{fig:fullfat_appendix}
\end{figure}

In section \ref{sec:experiment_fullfat} we evaluate GPEC's ability to combine uncertainty from the black-box decision boundary and the uncertainty estimate from BayesSHAP and SSV explainers. In Figure \ref{fig:fullfat_appendix} we extend this experiment to evaluate how well GPEC can capture the explainer uncertainty. We calculate the combined GPEC and explainer estimate using different numbers of approximation samples.

Both BayesSHAP and SSV depend on sampling to generate their explanations; having fewer samples increases the variance of their estimates. As we decrease the number of samples from 200 (Row A) to 5 (Row B) we would expect that the explainer uncertainty, and consequently the combined GPEC uncertainty, would increase. We see in Row $\Delta$ that the results follow our intuition; uncertainty increases for most of the plotted test points and uncertainty does not decrease for any points.

\subsection{Regularization Experiment: Standard Error} \label{app:addnl_results_regularization}
In Table \ref{table:regularize_var} and \ref{table:regularize_parameter}, we present the standard error measurements and parameters, respectively, for the regularization experiment.

\subsection{Additional Results for Uncertainty Visualization Experiment} \label{app:addnl_results_uncertainty_visualization}

\begin{table*}[t]
    %------------------------------------------
    % \setlength{\belowcaptionskip}{-15pt}

    \scriptsize
    % \small
    \setlength{\tabcolsep}{2.3pt}
    \renewcommand{\arraystretch}{1.1}
    \centering
    \begin{tabular}{ l c c c c c c c c c c c c c c c}
        %--------------
        \cline{0-15}
        \multicolumn{1}{|l|}{\emph{Dataset}} & 
        \multicolumn{3}{c|}{\textbf{Census}}  & 
        \multicolumn{3}{c|}{\textbf{Online Shoppers}} &
        \multicolumn{3}{c|}{\textbf{German Credit}} &
        \multicolumn{6}{c|}{\textbf{CIFAR10}} \\
        % \multicolumn{3}{c}{\textbf{ }} \\
        \cline{0-15}

        %--------------
        \multicolumn{1}{|l|}{\emph{Regularization}} & 
        \multicolumn{3}{c|}{$\gamma$}  & 
        \multicolumn{3}{c|}{$\gamma$} &
        \multicolumn{3}{c|}{$\gamma$} &
        \multicolumn{3}{c|}{$\ell_2$} &
        \multicolumn{3}{c|}{Softplus $\beta$} \\
        \cline{0-15}
        %--------------
        \multicolumn{1}{|l|}{\emph{Magnitude}} & 
        \multicolumn{1}{c}{$0$}  & 
        \multicolumn{1}{c}{$5$}  & 
        \multicolumn{1}{c|}{$10$}  & 
        \multicolumn{1}{c}{$0$}  & 
        \multicolumn{1}{c}{$5$}  & 
        \multicolumn{1}{c|}{$10$}  & 
        \multicolumn{1}{c}{$0$}  & 
        \multicolumn{1}{c}{$5$}  & 
        \multicolumn{1}{c|}{$10$} &
        \multicolumn{1}{c}{$0$}  & 
        \multicolumn{1}{c}{$1\textrm{e-5}$}  & 
        \multicolumn{1}{c|}{$10\textrm{e-5}$}  & 
        \multicolumn{1}{c}{$1.0$}  & 
        \multicolumn{1}{c}{$0.5$}  & 
        \multicolumn{1}{c|}{$0.25$} \\
        \cline{0-15} 
        %--------------
            
        \multicolumn{1}{|l|}{GPEC} &
            0.037 &
            0.037 &
            \multicolumn{1}{c|}{0.036}  &
            %---
            0.017 &
            0.013 &
            \multicolumn{1}{c|}{0.010}  &
            %---
            0.026 &
            0.030 &
            \multicolumn{1}{c|}{0.012}  & 
            %---
            2.8e-5 &
            2.8e-5 &
            \multicolumn{1}{c|}{2.8e-5}  & 
            %---
            2.5e-5 &
            2.5e-5 &
            \multicolumn{1}{c|}{2.5e-5}   

            \\

        %--------------
        \multicolumn{1}{|l|}{Naive-GP} &
            0.032 &
            0.032 &
            \multicolumn{1}{c|}{0.032}  &
            %---
            0.036 &
            0.036 &
            \multicolumn{1}{c|}{0.036}  &
            %---
            0.004 &
            0.001 &
            \multicolumn{1}{c|}{0.001} & 
            %---
            1.6e-3 &
            1.6e-3 &
            \multicolumn{1}{c|}{1.6e-3} & 
            %---
            1.6e-3 &
            1.6e-3 &
            \multicolumn{1}{c|}{1.6e-3}
            \\

        %--------------
        \multicolumn{1}{|l|}{BayesSHAP} &
            1.9e-4 &
            1.9e-4 &
            \multicolumn{1}{c|}{1.9e-4}  &
            %---
            1.9e-4 &
            1.9e-4 &
            \multicolumn{1}{c|}{1.9e-4}  &
            %---
            1.2e-4 &
            1.0e-4 &
            \multicolumn{1}{c|}{0.7e-4}  & 
            %---
            -- &
            -- &
            \multicolumn{1}{c|}{--}  & 
            %---
            -- &
            -- &
            \multicolumn{1}{c|}{--}   
            \\
            
        %--------------
        \multicolumn{1}{|l|}{BayesLIME} &
            0.001 &
            0.001 &
            \multicolumn{1}{c|}{0.001}  &
            %---
            0.002 &
            0.002 &
            \multicolumn{1}{c|}{0.002}  &
            %---
            0.001 &
            0.001 &
            \multicolumn{1}{c|}{0.001}  & 
            %---
            -- &
            -- &
            \multicolumn{1}{c|}{--}  & 
            %---
            -- &
            -- &
            \multicolumn{1}{c|}{--}   
            \\
            
        %--------------
        \multicolumn{1}{|l|}{CXPlain} &
            0.006 &
            0.006 &
            \multicolumn{1}{c|}{0.006}  &
            %---
            7.8e-5 &
            9.9e-5 &
            \multicolumn{1}{c|}{2.2e-4} &
            %---
            2.2e-6 &
            1.4e-6 &
            \multicolumn{1}{c|}{6.9e-6}  & 
            %---
            1.9e-7 &
            5.7e-7 &
            \multicolumn{1}{c|}{5.2e-7}  & 
            %---
            2.8e-8 &
            5.8e-8 &
            \multicolumn{1}{c|}{1.7e-8} 
            \\

        %--------------
        \cline{0-15}
        &
            &
            &
            &
            %---
            &
            &
            &
            %---
            &
            & 
            & 
            %---
            &
            & 
            & 
            %---
            &
            & 
            \\
        %--------------

        \cline{0-12}

        \multicolumn{1}{|l|}{\emph{Dataset}} & 
        \multicolumn{6}{c|}{\textbf{MNIST}}  & 
        \multicolumn{6}{c|}{\textbf{Fashion MNIST}}
         \\
        \cline{0-12}
        %--------------
        \multicolumn{1}{|l|}{\emph{Regularization}} & 
        \multicolumn{3}{c|}{$\ell_2$}  & 
        \multicolumn{3}{c|}{Softplus $\beta$}  & 
        \multicolumn{3}{c|}{$\ell_2$}  & 
        \multicolumn{3}{c|}{Softplus $\beta$} 
         \\
        \cline{0-12}
        %--------------
        \multicolumn{1}{|l|}{\emph{Magnitude}} & 
        \multicolumn{1}{c}{$0$}  & 
        \multicolumn{1}{c}{$1\textrm{e-5}$}  & 
        \multicolumn{1}{c|}{$10\textrm{e-5}$}  & 
        \multicolumn{1}{c}{$1.0$}  & 
        \multicolumn{1}{c}{$0.5$}  & 
        \multicolumn{1}{c|}{$0.25$}  & 
        \multicolumn{1}{c}{$0$}  & 
        \multicolumn{1}{c}{$1\textrm{e-5}$}  & 
        \multicolumn{1}{c|}{$10\textrm{e-5}$}  & 
        \multicolumn{1}{c}{$1.0$}  & 
        \multicolumn{1}{c}{$0.5$}  & 
        \multicolumn{1}{c|}{$0.25$} 
         \\
        \cline{0-12}
        %--------------

        \multicolumn{1}{|l|}{GPEC} &
            3.2e-5 &
            3.1e-5 &
            \multicolumn{1}{c|}{3.1e-5}  & 
            %---
            2.2e-6 &
            2.2e-6 &
            \multicolumn{1}{c|}{2.2e-6}  & 
            %---
            8.3e-5 &
            8.3e-5 &
            \multicolumn{1}{c|}{8.1e-5}  & 
            %---
            2.1e-6 &
            1.9e-6 &
            \multicolumn{1}{c|}{1.9e-6}  & 

            \\

        %--------------
        \multicolumn{1}{|l|}{Naive-GP} &
            1.6e-7 &
            1.6e-7 &
            \multicolumn{1}{c|}{1.6e-7}  &
            %---
            2.3e-7 &
            2.3e-7 &
            \multicolumn{1}{c|}{2.2e-7}  &
            %---
            3.6e-7 &
            3.6e-7 &
            \multicolumn{1}{c|}{3.6e-7}  &
            %---
            3.6e-7 &
            3.5e-7 &
            \multicolumn{1}{c|}{3.6e-7}  &

            \\

        %--------------
        \multicolumn{1}{|l|}{BayesSHAP} &
            4.9e-4 &
            3.1e-4 &
            \multicolumn{1}{c|}{2.1e-4}  &
            %---
            4.5e-4 &
            4.2e-4 &
            \multicolumn{1}{c|}{4.2e-4}  &
            %---
            1.3e-3&
            0.5e-3 &
            \multicolumn{1}{c|}{0.2e-3}  &
            %---
            0.002 &
            0.002 &
            \multicolumn{1}{c|}{0.002} 

            \\
            
        %--------------
        \multicolumn{1}{|l|}{BayesLIME} &
            0.009 &
            0.009 &
            \multicolumn{1}{c|}{0.003}  &
            %---
            0.005 &
            0.005 &
            \multicolumn{1}{c|}{0.005}  &
            %---
            0.067 &
            0.038 &
            \multicolumn{1}{c|}{0.009}  &
            %---
            0.010 &
            0.011 &
            \multicolumn{1}{c|}{0.011} 

            \\
            
        %--------------
        \multicolumn{1}{|l|}{CXPlain} &
            0.1$\textrm{e-5}$ &
            5.0$\textrm{e-5}$ &
            \multicolumn{1}{c|}{8.6 $\textrm{e-5}$}  &
            %---
            5.3$\textrm{e-5}$ &
            8.0$\textrm{e-5}$ &
            \multicolumn{1}{c|}{5.4$\textrm{e-5}$}  &
            %---
            7.2$\textrm{e-5}$ &
            4.8$\textrm{e-5}$ &
            \multicolumn{1}{c|}{6.2 $\textrm{e-5}$}  &
            %---
            9.0$\textrm{e-5}$ &
            6.6$\textrm{e-5}$ &
            \multicolumn{1}{c|}{9.6$\textrm{e-5}$} 

            \\
        %--------------
    
        \cline{0-12}
    \end{tabular}
    \caption{Standard Error for results in Table \ref{table:regularize}}
    \label{table:regularize_var}
    
    \end{table*}

\begin{table*}[t]
    %------------------------------------------

    \scriptsize
    % \small
    % \setlength{\tabcolsep}{2.3pt}
    \renewcommand{\arraystretch}{1.1}
    \centering
    \begin{tabular}{|l c c|} 
    \hline
     \textbf{Dataset} & $\lambda$  & $\rho$ \\ 
     \hline
     Census & $1.0$ & $0.1$\\ 
     Online Shoppers & $1.0$ & $0.1$\\ 
     German Credit & $1.0$ & $0.1$\\ 
     MNIST & $1.0$ & $0.01$\\ 
     f-MNIST & $1.0$ & $0.01$\\ 
     CIFAR10 & $1.0$ & $0.05$\\ 
     \hline
    \end{tabular}
    \caption{GPEC Parameters for results in Table \ref{table:regularize}}
    \label{table:regularize_parameter}
    
    \end{table*}

\begin{figure}[t]
   \begin{center}
   \includegraphics[width=1.0\linewidth]{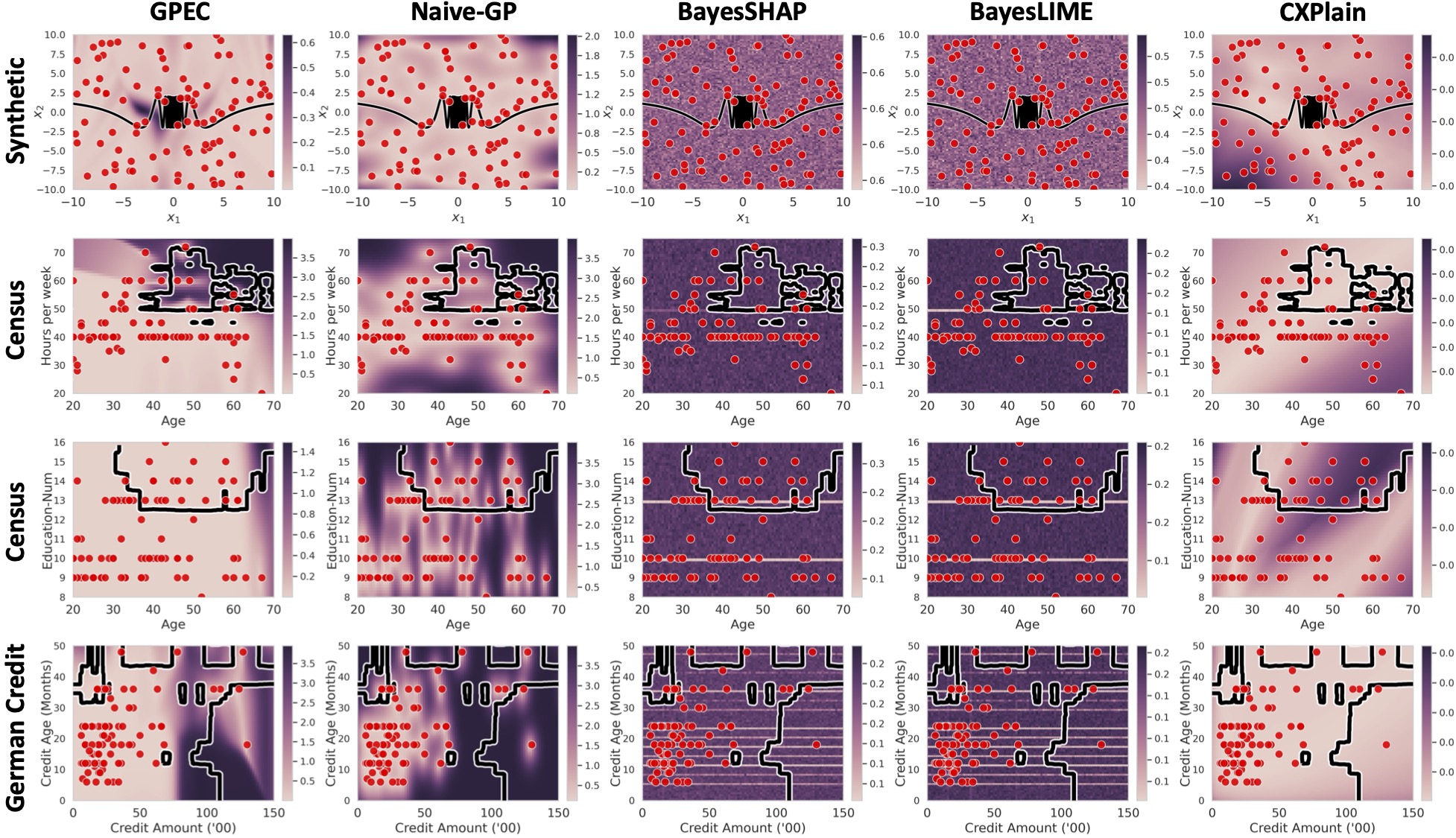}
   \end{center}
   \caption{Complement to Figure \ref{fig:uncertaintyfigure}. Visualization of estimated explanation uncertainty where the heatmap represents level of uncertainty for the feature on the \textbf{y-axis}.}
   \label{fig:uncertaintyfigure_feat1}
\end{figure}

In Figure \ref{fig:uncertaintyfigure_feat1} we visualize the estimated explanation uncertainty as a heatmap for a grid of explanations. The generated plots only visualize the uncertainty for the feature on the x-axis. Due to space constraints, we list the results for the y-axis feature in the appendix, in Figure \ref{fig:uncertaintyfigure_feat1}. We can see that the results are in line with those from the x-axis figure.

\end{document}